\documentclass[preprint,authoryear,11pt]{elsarticle}

\usepackage{graphicx}

\usepackage{amssymb}

\usepackage[top=3.5cm,left=3.5cm,right=3.5cm,bottom=3.5cm]{geometry}

\usepackage{enumitem} 
  \setlist{itemsep=1ex plus0.2ex, leftmargin=*, align=left}

\makeatletter
\newcommand{\labitem}[2]{%
\def\@itemlabel{\textbf{#1}}
\item
\def\@currentlabel{#1}\label{#2}}
\makeatother

\makeatletter
\newcommand{\headingitem}[1]{%
\vspace{0.3cm}
\def\@itemlabel{\textbf{#1}}
\item
\def\@currentlabel{#1}
\addtocounter{enumi}{-1}
}
\makeatother

\usepackage{csvsimple}
\usepackage{multirow}
\usepackage{lipsum}
\usepackage{afterpage}

\usepackage{etoolbox}
\robustify\bfseries

\usepackage{eurosym}
\usepackage{siunitx}

    \DeclareSIUnit\eur{\officialeuro}
    \DeclareSIUnit\M{M}
    \DeclareSIUnit\k{k}

  \def\sym#1{\ifmmode^{#1}\else\(^{#1}\)\fi}

\usepackage{setspace}
\usepackage{csquotes}
\usepackage{amsmath}
\usepackage{bm}

\usepackage{xspace}
	\newcommand\ie{i.\,e.\xspace}
	\newcommand\eg{e.\,g.\xspace}

	\newcommand\etc{etc.\xspace}
  
	\newcommand\cf{cf.\xspace}

	\newcommand\US{U.\,S.\xspace}
	\newcommand\EU{E.\,U.\xspace}

\usepackage[amsmath,hyperref,framed]{ntheorem} 
  \theoremstyle{plain}
     
  \theoremstyle{nonumberplain}
    \theoremseparator{.}
    \theoremheaderfont{\bfseries}
    \theorembodyfont{\normalfont}
    \theoremsymbol{$\blacksquare$}
      \RequirePackage{amssymb}

      \makeatletter
    \let\copy@theorem@headerfont=\theorem@headerfont
    \newcommand{\my@theorem@headerfont}{%
        \boldmath\copy@theorem@headerfont\unboldmath
      }
    \let\theorem@headerfont=\my@theorem@headerfont
      \makeatother

\theoremstyle{nonumberplain}
\theoremseparator{.}
\usepackage[%
  sort&compress
]{cleveref}
\usepackage{url}

\usepackage{isomath}

  \newcommand{\norm}[1]{\left\lVert#1\right\rVert}

\usepackage{algorithmic}        

    \crefname{ALC@unique}{step}{steps}
    \Crefname{ALC@unique}{Step}{Steps}
    \crefname{ALC@line}{step}{steps}
    \Crefname{ALC@line}{Step}{Steps}

\usepackage[usenames,dvipsnames]{xcolor}

\usepackage{colortbl}
\usepackage{tabularx}
\usepackage{booktabs}
\usepackage{collcell}

\usepackage{longtable}

\usepackage{ragged2e}
  
\newcommand{\PreserveBackslash}[1]{\let\temp=\\#1\let\\=\temp}
\newcolumntype{v}[1]{>{\PreserveBackslash\RaggedRight\hspace{0pt}}p{#1}}

  \usepackage{adjustbox}
\newcolumntype{Q}[2]{%
    >{\adjustbox{angle=#1,lap=\width-(#2)}\bgroup}%
    l%
    <{\egroup}%
}

\newcommand{\mcellt}[2][c]{%
  \begin{tabular}[t]{@{}#1@{}}#2\end{tabular}}

\usepackage[
    colorinlistoftodos,
    textsize=footnotesize,
        ]{todonotes}

\makeatletter
    \renewcommand{\fps@figure}{htb}         
    \renewcommand{\fps@table}{htb}         
\makeatother 

\usepackage{float}
\usepackage{rotating}

\journal{arXiv}

\begin{document}

\begin{frontmatter}



\title{News-based forecasts of macroeconomic indicators:\\
A semantic path model for interpretable predictions}


\author[ETH,Freiburg]{Stefan Feuerriegel\corref{cor1}}
\ead{sfeuerriegel@ethz.ch}

\author[Freiburg]{Julius Gordon}
\ead{juliusgordon89@gmail.com}

\address[ETH]{ETH Zurich, Weinbergstr. 56/58, 8092 Zurich, Switzerland}
\address[Freiburg]{Chair for Informaton Systems Research, University of Freiburg, Platz der Alten Synagoge, 79098 Freiburg, Germany}

\cortext[cor1]{Corresponding author.}

\begin{abstract}
The macroeconomic climate influences operations with regard to, e.g., raw material prices, financing, supply chain utilization and demand quotas. In order to adapt to the economic environment, decision-makers across the public and private sectors require accurate forecasts of the economic outlook. Existing predictive frameworks base their forecasts primarily on time series analysis, as well as the judgments of experts. As a consequence, current approaches are often biased and prone to error. In order to reduce forecast errors, this paper presents an innovative methodology that extends lag variables with unstructured data in the form of financial news: (1) we apply a variety of models from machine learning to word counts as a high-dimensional input. However, this approach suffers from low interpretability and overfitting, motivating the following remedies. (2) We follow the intuition that the economic climate is driven by general sentiments and suggest a projection of words onto latent semantic structures as a means of feature engineering. (3) We propose a semantic path model, together with estimation technique based on regularization, in order to yield full interpretability of the forecasts. We demonstrate the predictive performance of our approach by utilizing 80,813 ad hoc announcements in order to make long-term forecasts of up to 24 months ahead regarding key macroeconomic indicators. Back-testing reveals a considerable reduction in forecast errors.
\end{abstract}


\end{frontmatter}

\section{Introduction}


Macroeconomic developments, such as cyclic downturns or the economic circumstances associated with the \US subprime crisis,  affect firm operations in multiple ways and represent direct challenges to management \citep[e.\,g.][]{Demyanyk.2010,Goudie.1982}. Examples include changes in the price of goods and raw materials, as well as the impact on overall demand, supply chain utilization and even operational processes \citep{Xu.2017}. Therefore, firms are interested in foreseeing the future economic climate in order to manage operations accordingly and hedge potential risks. In this context, operational research~(OR) has a long tradition of addressing such risks \citep{Demyanyk.2010}. Our discipline has thus contributed to anticipating a variety of developments at a macroeconomic level, including financial distress \citep{Geng.2015}, liquidity risks \citep{Shaik.2015}, credit risks \citep{Akkoc.2012,Desai.1996}, financial crises \citep{Demyanyk.2010,Huang.2017}, currency crises \citep{Sevim.2014} and bankruptcy \citep{DuJardin.2015,McKee.2002,Sun.2007}, especially in the financial sector \citep{Tam.1992}


Future expectations regarding the macroeconomic environment play a critical role in the decision-making process for many organizations \citep{Xu.2017}. Hence, decision-makers across all sectors must analyze the current economic environment and form accurate expectations about future economic trends in order to support the operational strategy of organization and long-term management. As a result, macroeconomic variables, and the accurate prediction thereof, form the basis for a wide array of OR models \citep[e.\,g.][]{Calabrese.2017,Fethi.2010,Gutierrez.2012}.


The importance of accurate long-term forecasts for firm operations has driven the extensive amount of research conducted with respect to macroeconomic predictions. Specific examples from the OR domain include short-term predictions of asset-related values, including government bonds \citep{Tay.2001} and stock indices \citep[e.\,g.][]{Huang.2005,Kung.2008,Oztekin.2016}. Further research focuses on forecasting macroeconomic indicators of single countries \citep[e.\,g.][]{Mahmoud.1990} or the relationship between countries \citep{Sermpinis.2013}. Other works propose agent-based simulations to study the behavior of human forecasters \citep{Bovi.2016}. 


Previous efforts at forecasting macroeconomic indicators have made use of various input features and methodologies. Historic time series data is a staple input for macroeconomic forecasting, and has been applied to make both short- and long-term predictions \citep[e.\,g.][]{Jansen.2016,Mahmoud.1990,Sermpinis.2013}. A prevalent alternative is the subjective judgments of professional forecasters \citep{Matsypura.2018} such as those used by the European Central Bank. However, both time series models and subjective forecasts  suffer from severe prediction errors \citep{Jansen.2016}, possibly because they cannot benefit from predictors that better measure the state and outlook of the economy. 


It is plausible that macroeconomic forecasting could also benefit from the advances of the big data era \citep{Mortenson.2015} and potential improvements in predictive modeling. Especially unstructured data, such as news, promises further insights into global systems and the heterogeneous trends and events that occur within them. Hence, news represents a powerful source of information for financial forecasting and is thus likely to serve as an intriguing -- yet currently rare -- predictor for macroeconomic forecasts. 


Our research therefore aims to improve macroeconomic forecasts using financial disclosures. Financial news, in particular, conveys rich information about expected firm performance that often goes beyond pure numeric data \citep[e.\,g.][]{Tsai.2017}. Text mining of financial news has been successfully used in conjunction with predictive analytics to infer profitable short-term investment decisions in capital markets \citep{Nassirtoussi.2014, Feuerriegel.2016}. However, it is unclear whether financial filings -- which should essentially mirror the current health of the firm -- also provide long-term prognostic capacity.   



To make long-term predictions of macroeconomic indicators, we construct an approach as follows: we use financial news and insert corresponding linguistic features into various machine learning methods. Our specific setting represents a major difference from previous news-based forecasts, in the sense that the outcome variable is reported in monthly or quarterly resolution. This results in fewer observations and thus extremely high-dimensional predictor matrices with severe risks of overfitting. To overcome this challenge, we develop a form of feature engineering based on semantic projections and, on top of that, propose a semantic path model that fulfills the demand of practitioners by being fully interpretable. Afterwards, we conduct an out-of-sample evaluation of the predictive performance. The results demonstrate that our approach is superior in terms of relative performance when compared to common time series models that serve as a benchmark.

The remainder of this paper is structured as follows. \Cref{sec:background} reviews related research on macroeconomic forecasting in order to outline the aforementioned research gap. As a remedy, \Cref{sec:methods} explains our methodology for making news-based macroeconomic predictions, with a particular focus on the proposed projections to latent semantic structures. \Cref{sec:datasets} reports the datasets, based on which \Cref{sec:results} compares the predictive performance of both traditional time series models and news-based forecasts. A discussion of managerial implications follows in \Cref{sec:discussion}, while \Cref{sec:conclusion} concludes with an outlook on future research.

\section{Background} 
\label{sec:background}

\subsection{Predictive models for macroeconomic forecasts}


Beyond human predictions \citep{Matsypura.2018}, common approaches to macroeconomic forecasting include various time series models \citep{Allen.2006}, for instance, auto-regressive moving-average~(ARMA) models and their multivariate variations, such as vector autoregression and vector error correction models. However, their use entails certain challenges, since they must be adapted to cope with structural changes in the underlying systems \citep{Litterman.1986}. Nevertheless, time series models still appear to be the standard benchmark against which the predictive performance of alternative approaches is measured \citep{Gooijer.2006}. 

Since vector autoregression and its variants are limited to a few variables, researchers have proposed alternative models, such as Bayesian vector autoregression~(BVAR). The BVAR accounts more accurately for uncertainty regarding the structure of the economy and is utilized to handle instances with dozens of relevant variables \citep{Litterman.1986}. Its use for forecasts requires extensive computational resources, especially when dealing with many predictors \citep{Carriero.2015,Litterman.1986}. 


Bridge equations have been a suggested as an alternative means of incorporating relevant explanatory variables into predictive models. This approach is a form of linear regression that time-aggregates economic time series to bridge the information gap between low and high frequency indicators. It is a widely used method for forecasting macroeconomic variables \citep{Jansen.2016}. However, these forecasts are generally limited to one or two periods ahead. 

A different concept is represented by leading indicators. These are supposed to provide prognostic capabilities for future changes in the economic outlook. Examples include major stock indices, the Ifo Business Climate Index from Germany, or the University of Michigan Consumer Sentiment Index for the United States. These indices can yield important information regarding future cyclical fluctuations of the general economy \citep{Allen.2006} and they have thus been used to forecast imminent turning points in the business cycle \citep{Layton.2007}. Predictive models using leading indicators have been applied, for instance, in single variable ARMA models, multivariate models, diffusion indices and BVAR \citep{Allen.2006,Stock.2002}. 
 However, it is not clear whether these indices have a prognostic potential for future developments. 

Since previous approaches to macroeconomic forecasting have primarily focused on using various forms of economic data as dynamic predictors, the use of qualitative or unstructured data, such as financial news, deserves the attention of researchers. However, the inclusion of larger numbers of predictors in traditional time series models can prove to be problematic, as such models typically struggle with such settings \citep{Jansen.2016, Litterman.1986}. The above literature review thus motivates our choice of models: we specifically decided to utilize different machine learning models that can cope with non-linear relationships and large numbers of textual features. 


\subsection{Predictive text mining with financial news}


News-based predictions of macroeconomic indicators constitute an innovative area of research. Due to the scarcity of related publications, we decided to extend our literature overview and also include news-based forecasts of stock returns in addition to macroeconomic variables.

Text mining of financial news have been successfully used in conjunction with predictive analytics to infer profitable short-term investment decisions in capital markets \citep{Feuerriegel.2016}. Previous approaches have largely focused on predicting the response to financial news in the form of \emph{short-term} reactions of \emph{stock markets} \citep{Nassirtoussi.2014}, whereas we develop a forecasting methodology for the general \emph{macroeconomic environment} over \emph{long-term forecast horizons}. Text mining of financial news for predictions in capital markets has been extensively studied in recent years; for a comprehensive review we refer to \citep{Nassirtoussi.2014}.



Financial news contains fundamental and qualitative information that influences the expectations of market participants, and has thus driven the field of market prediction. A broad range of textual materials has been in the course of recent research, including newspaper articles from media sources and regulated filings, such as ad~hoc announcements, 8-K filings and annual reports \citep{Nassirtoussi.2014}. The common feature tying these previous efforts together is the focus on making short-term, usually same-day, predictions. 

The challenge of working with textual data, such as financial news, lies in how to efficiently transform it into a machine-friendly form and then apply a classifier to it. Hence, text mining approaches commonly follow a similar procedure with the goal of extracting and selecting relevant features that best represent the original text \citep{Nassirtoussi.2014}. In this regard, the bag-of-words approach breaks up the free-running text into single words (or $n$-grams) which can then be used as features. Afterwards, feature selection (sometimes joined with dimensionality reduction) aims to reduce the set of features while keeping relevant information. Common approaches in financial text mining \citep{Nassirtoussi.2014} include frequency-based statistics, \ie term weighting. This then presents the input to machine learning classifiers. We draw upon the aforementioned approach utilizing high-dimensional input matrices, which results in one of our three text-based forecasting routines. 

\section{Methods} 
\label{sec:methods}

\subsection{Overview}



This section introduces the research framework for forecasting macroeconomic indicators based on financial news. This task entails two critical challenges: feature engineering and interpretability.

First, the input matrix is high-dimensional as the number of predictors exceeds the number of observations. For instance, the macroeconomic indicators with quarterly resolution are predicted based on a term-document matrix with 536 terms, while a single fold during cross-validation might consist of only 71 samples. We address this issue through unsupervised techniques for dimension reduction and, in addition, pioneer in replacing the statistical approaches to feature generation \citep[e.\,g.][]{Liang.2016} with a supervised technique that incorporates domain-specific knowledge. More specifically, it aggregates terms according to a semantic classification with projections onto corresponding constructs, thereby yielding a low-dimensional input with only a couple of predictors.

Second, machine learning is often regarded as a black-box mechanism, whereas practitioners from operational management demand the potential for interpretation as a means to validate the computer-based forecasts with their own beliefs and expectations, as well as to gain trust in the automated decision support. Hence, we are specifically interested in models that attain a beneficial trade-off between predictive performance and explanatory power. As a remedy, we propose a semantic path model that can offer full accountability of the forecasts but yields a performance comparable to the best black-box models from machine learning. Here we follow the intuition that the economic climate is linked to the perception (\ie sentiments) of the involved stakeholders. Given a meaningful choice of dimensions, we can later decompose the overall forecasts in terms of the underlying constructs.


We adapt to the two aforementioned challenges by suggesting a variety of predictive methods for our forecasting task. The underlying approach for text-based forecasts consists of preprocessing of financial text, including feature extraction (\cf online appendix); optional feature engineering, in which we project individual words onto latent semantic structures; model training and selection; and multi-step ahead predictions using a test set. The individual approaches can be grouped as follows: 
\begin{enumerate}[start=0]
\item \emph{Baseline: Time series models.} These build upon auto-regressive lags and combine them in linear and non-linear fashion.
\item \emph{Machine learning models with high-dimensional input}. A series of machine learning models is tested with high-dimensional input as a comparison. The models are later fed with an input matrix that is (optionally) subject to unsupervised dimension reduction techniques.
\item \emph{Models with semantic features.} We develop a supervised technique for feature engineering. This approach involves projections of terms onto semantic constructs that then serve as a low-dimensional feature space.
\item \emph{Semantic path model.} Similar to the semantic features, we develop a path model that operationalizes the semantic classification of terms into different categories relevant in a macroeconomic context in order to obtain interpretable forecasts. To circumvent overfitting from collinear constructs, we develop a novel estimation technique with an additional regularization.
\end{enumerate}
We note that each of the text-based models is later trained in different variants, \ie the models are optionally augmented by the same lags as the time series baseline. 


\subsection{Baseline: Time series models}

In order to evaluate the predictive performance of our forecasting methodology, we employ a number of benchmark models that forecast the macroeconomic time series purely on the basis of historic data points. These models include an auto-regressive~(AR) process based on $l$ lagged values. Formally, let $i$ denote the current time period. We then predict the value in time step $i+h$, \ie that is $h$ steps ahead. This results in
\begin{equation} 
\label{eq:AR}
Y_{i+h} = {\alpha} + {\beta}_1 Y_{i-1} + \ldots + {\beta}_l Y_{i-l} + {\varepsilon}_i ,
\end{equation}
with coefficients $\alpha, \beta_1, \ldots, \beta_l$, inputs $Y_{i-1}, \ldots, Y_{i-l}$ and where ${Y}_{i+h}$ gives the predicted indicator. In addition, we incorporate non-linear relationships by applying a random forest~(RF) to historic values for each of the macroeconomic indicators. This machine learning model frequently achieves a high predictive power with little need for tuning \citep{Hastie.2013}. 

For both AR and RF models, we experiment with different specifications that vary in their lag order: on the one hand, we utilize only \num{1} lag and, on the other hand, all lags of order 1 to 6. By testing both variants, we also adhere to the selection made in previous literature \citep[e.\,g.][]{Litterman.1986,Stock.2002}. We specifically decided to predict the raw values of each of the macroeconomic indicators, instead of applying additional transformation. This choice stems from the need of practitioners who require clear expectations in common and interpretable scales. As a comparison, we also predict a first-difference term as part of our evaluation.

\subsection{Models for textual predictors}

All subsequent models involving textual predictors are estimated in three different variants: on the one hand, we forecast macroeconomic indicators based purely on the narrative materials, excluding the history of the outcome variable. On the other hand, we follow the above time series models and augment the predictor matrix by 1 and 6 auto-regressive lags of the to-be-predicted macroeconomic indicator. Thus the model can further learn from additional seasonal and trend characteristics. 

\subsubsection{Machine learning models for high-dimensional input}


As noted earlier, the document-term matrices are wide, since the number of features~(words) far exceeds the number of observations. We thus choose machine learning models that are known for their performance with wide data sets, as they reduce the risk of overfitting via feature selection, regularization and dimension reduction \citep{Hastie.2013}. These are as follows: least squares absolute shrinkage operator~(LASSO), ridge regression, elastic net~(ENET), gradient boosting~(GBM), principal component regression~(PCR), random forest~(RF) and partial least squares regression~(PLS-R). 


Following common conventions, the above models are estimated based on the term-document matrix.  The entries are scaled by tf-idf weighting in order to reflect how frequently a term appears in a document in the corpus. Beyond that, the experiments are repeated with different unsupervised transformations for dimension reduction that are applied to the input $X$ as part of a feature engineering step. More precisely, we apply a principal component analysis~(PCA) and compute a latent semantic analysis~(LSA), which are then fed into the different machine learning models. Altogether, we obtain a total of 21 different models (\ie 7 classifiers, each with raw tf-idf, as well as 2 unsupervised dimension reduction techniques).

\subsubsection{Models with semantic features}

We now present an alternative approach to feature generation. More precisely, we replace the purely statistical approach behind unsupervised transformations for dimension reduction (\ie PCA and LSA) with a supervised routine that encodes domain-specific knowledge.

Our approach assumes that the overall macroeconomic climate is driven by the perceptions (or sentiments) of individual stakeholders. For instance, a construct \textquote{macroeconomic uncertainty} could be indicated by a term \textquote{risk}, while \textquote{increase} links to a construct \textquote{positive outlook}. Hence, we identify a new set of features $z_1, \ldots, z_m$ which represent such meaningful constructs (\ie perceptions). Each construct $z_i$ is represented by a linear combination of terms $x_1, \ldots, x_n$ or, more specifically, a subset $\mathcal{I}_i \subseteq \left\{ 1, \ldots, n \right\}$ thereof; \ie $z_i = \sum_{j\in\mathcal{I}_i} \phi_{ij} x_j$ with coefficients $\phi_{ij}$. The subset then incorporates our knowledge with regard to how terms correspond to the different constructs. Here merely a categorization $\mathcal{I}_i$ of words is needed in the form of a simple wordlist that neither has weights nor is mutually exclusive. As a result, the high-dimensional input matrix is projected onto a low-dimensional feature space based on the the pre-specified semantic structure.

The weights $\phi_{ij}$ are computed in a supervised manner by making use of the response $Y$. That is, the weights $\phi_{ij}$ are chosen such that the $z_i$ explain as much of the covariance between $X$ and $Y$ as possible. Mathematically, the computation draws upon the existing routine for projections on latent constructs \citep{Hastie.2013}. This algorithm computes the coefficients for the each construct $z_i$ similar to least squares, \ie $\min_{\phi_{ij}, \ldots, \phi_{im}} \; \norm{Y - \phi X}_2^2$, with the difference that it involves the orthogonalization step from path modeling, as terms can appear in more than one construct \citep{Tenenhaus.2005}. This highlights the differences from traditional PLS-R as a predictive model, in which the number of constructs $m$ is subjected to tuning and \emph{all} terms map on \emph{all} features. Conversely, our semantic features impose a structure where only a subset of terms maps to each feature, thereby incorporating domain knowledge.

{We additionally utilize a Bayesian neural network as the underlying predictive model in order to infer $Y$ from $z_i$, thereby essentially yielding a universal structure model, USM for short \citep{Buckler.2008}. In fact, universal structure modeling has found to be effective in multiple applications from the fields of decision support and operations research \citep[e.\,g.][]{Oztekin.2011,Turkyilmaz.2013,Turkyilmaz.2016}.}

\subsubsection{Semantic path model}



\begin{figure}
\centering
\includegraphics[width=.85\textwidth]{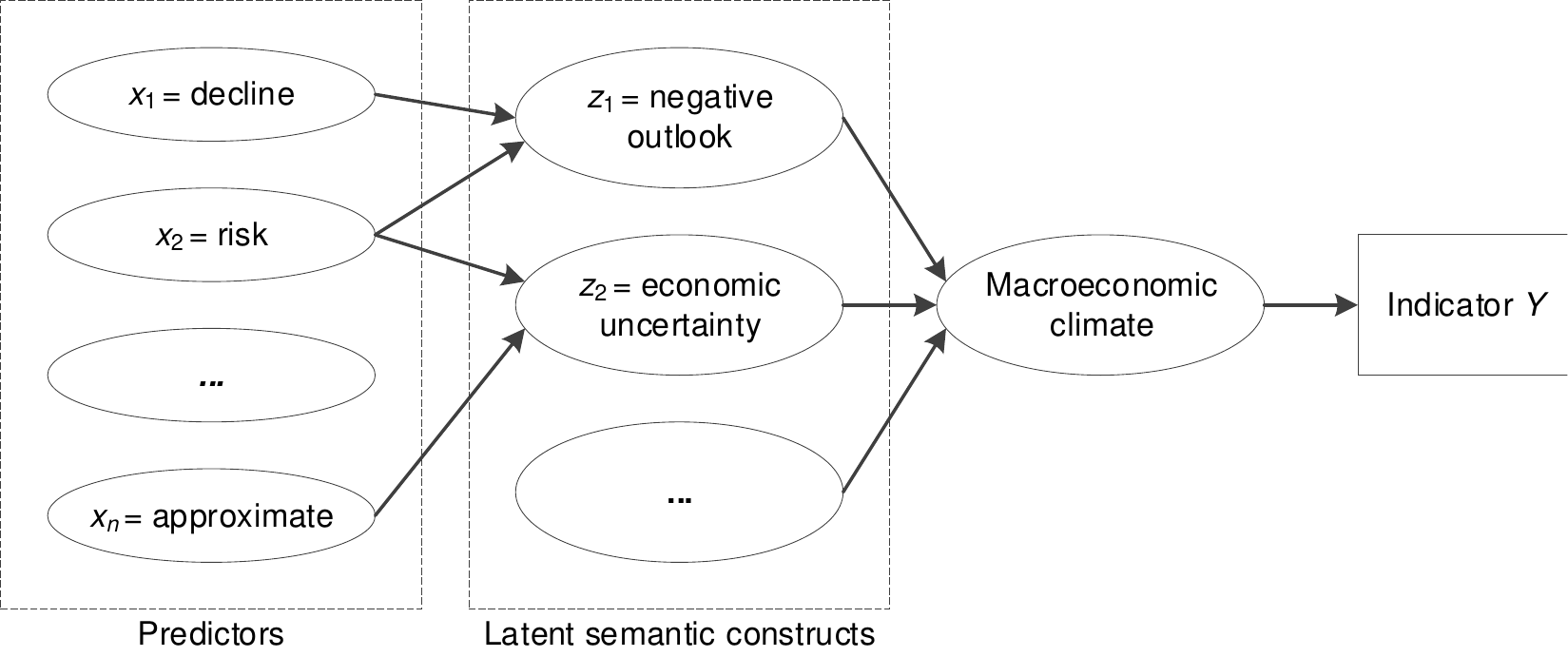}
\caption{Illustrative depiction of our semantic path model. The dictionary-based assignment of words to semantic categories is shown by the presence of arrow-shaped paths. Here projections map the high-dimensional predictors onto latent semantic constructs.}
\label{fig:PLS}
\end{figure}

We now present the semantic path model that also involve projections onto low-dimensional feature spaces. These again build upon a meaningful representation of different perceptions of the macroeconomic climate as latent constructs. However, this approach attains full interpretability as the underlying combination construct is linear. \Cref{fig:PLS} illustrates an example in which terms are mapped onto two constructs -- \emph{negative outlook} and \emph{economic uncertainty} -- which then yield the final prediction. {Such path modeling techniques are common in the social sciences \citep[cf.][]{AguirreUrreta.2014,Aubert.1996,McIntosh.2014}, where the constructs quantify different aspects of human behavior \citep{Tenenhaus.2005,Rigdon.2010}.} However, their adaptation to semantic structures for linguistic materials reveals an innovative application with considerable potential for interpreting language-based predictions.

Mathematically, the model builds upon the prescribed paths of which terms $\mathcal{I}_i \subseteq \left\{ 1, \ldots, n \right\}$ relate to construct $i$. Each construct is then computed by a linear combination $z_i = \sum_{j\in\mathcal{I}_i} \phi_{ij} x_j$, \ie a projection of words onto latent semantic constructs. Then, the final prediction is an additional linear model $\tilde{Y} = \psi_0 + \sum_{i=1}^{m} \psi_i \, z_i$. Evidently, one can directly interpret the forecast and decompose its value into individual components $\psi_1 z_1$, $\psi_2 z_2$, \etc Hence, decision-makers can assess the underlying reasoning of the model and evaluate the contribution of the different semantic categories to the overall outcome.

The above model can be estimated based on the algorithms for path modeling based on projections to latent structures \citep{Tenenhaus.2005}, which we denote in the following:
\begin{algorithmic}[1]
\footnotesize
\STATE Center $X$ and $Y$
\STATE Initialize outer weights $\phi_{ij}$ with arbitrary values (\eg $\phi_{ij} =1$)
\REPEAT
\STATE Compute the external approximation of latent constructs $z_i = \sum_{j\in\mathcal{I}_i} \phi_{ij} x_j$
\STATE Obtain inner weights and approximation of constructs
\STATE Calculate new outer weights $\phi_{ij}$
\UNTIL{convergence of outer weights $\phi_{ij}$}
\STATE Obtain inner weights, \ie path coefficients $\psi_i$, by ordinary least squares
\begin{equation}
\min_{\psi_0, \ldots, \psi_m} \norm{\tilde{Y} - \psi_0 - \sum_{i=1}^{m} \psi_i \, z_i}_2^2
\end{equation}
\end{algorithmic}

Beyond the na{\"i}ve estimation approach to path modeling based on projection to latent structures \citep{Tenenhaus.2005}, we propose an alternative strategy: we argue that the potentially collinear relationships between the different constructs could lead to a low capacity for generalization and thus suboptimal forecasts. This differs from the majority of path models in social sciences, where constructs represent distinct concepts \citep[e.\,g.][]{Aubert.1996} and thus mostly orthogonal vectors  \citep{Tenenhaus.2005}. However, in our case, a high negativity is likely to correspond to low positivity and vice versa. Hence, this can be addressed by replacing the ordinary least squares~(OLS) estimator from the inner model in path modeling with regularized variants, namely, LASSO, ridge regression and elastic net. This results in the general form
\begin{equation}
\min_{\psi_0, \ldots, \psi_m} \norm{\tilde{Y} - \psi_0 - \sum_{i=1}^{m} \psi_i \, z_i}_2^2 + \alpha_1 \, \norm{\psi}_1  + \alpha_2 \, \norm{\psi}_2^2,
\end{equation}
where the coefficients $\alpha_1$ and $\alpha_2$ control the amount of shrinkage and are subject to hyperparameter tuning. This merely affects the magnitude of the coefficients $\psi_i$ but maintains the linear relationship that lends itself to direct interpretation. In fact, the LASSO can set coefficients directly to zero and, accordingly, could even perform an implicit selection of relevant constructs. Closest to this are different approaches for shrinkage or the inclusion of sparsity in partial least squares regression \citep[e.\,g.][]{Chun.2010,LeCao.2008}, but, to the best of our knowledge, penalization has not yet been incorporated in path modeling.

{The semantic path model and PLS-R share similarities in terms of that both implement projections to latent structures. However, a clear difference becomes evident in how these mappings are defined. PLS-R follows a \emph{data-driven} approach where all terms map onto a number of latent constructs that is subject to tuning. Thereby, the latent constructs loose their ability of interpretation and collinearity among features -- such as in our high-dimensional setting -- could impede generalizability. In contrast, our semantic path model focuses on a \emph{knowledge-driven} methodology as it directly encodes a structure that lends to interpretation. The numerical results later indicate the potential performance improvements in our setting with extremely high-dimensional predictor matrices. Moreover, the proposed use of additional regularization can further bolster interpretation from more parsimonious structures and shrink the influence of collinear constructs, thereby reducing the risk of overfitting.}


\subsection{Model tuning}

All of the previous models from both benchmarks and news-based predictors were tuned by following the default parameter search grid from \citet{Kuhn.2008}. However, time series data does not lend itself to traditional random sampling used for cross-validation and we instead utilize a time-slice cross-validation with ten folds. This applies a rolling forecast origin, where the sample used for calibration contains only observations that occurred prior to the observations that form the validation set. The process is then repeated by rolling the origin of the validation set forward, thus incorporating the previous observations into the estimation procedure. The model is then retested using the next window of observations. Afterwards, the best configuration of hyperparameters is retained, based on which the final model is fitted to the complete training data. All models are tuned with the goal of minimizing the root mean squared errors. All models preprocess the predictors by scaling and centering to zero mean and a variance of one. 

\section{Datasets}
\label{sec:datasets}

\subsection{Financial news}

To make long-term predictions of macroeconomic indicators, we use financial news in the form of ad~hoc announcements. Ad~hoc announcements represent regulatory filings that are obligatory for all firms listed on the German stock exchange (\cf German Securities Trade Act for details). This dataset offers a number of advantages. First, German regulations require listed companies to immediately publish any information which could potentially impact the price of their stock via this type of announcement. Second, the content of the announcement is authorized by the executive board and finally checked by the Federal Financial Supervisory Authority. The tight regulations reduce the likelihood of non-relevant content. Third, standardized distribution channels guarantee real-time access to disclosures. This allows one to derive a live picture of the industrial sector, since corporations self-report their current performance but also incorporate statements regarding their optimism, future expectations and legal risks. Altogether, we possess a rich source of information to help sense -- potentially early -- economic developments from a production-based view. Ad~hoc announcements have been subject to extensive research revealing the benefits of their use in both practice and research, for instance, with regard to short-term investment decisions and risk predictions \citep[e.\,g.][]{Kraus.2017}.	

Our corpus is composed of all 80,813 ad~hoc announcements which we collected from the website of the EQS Group, the leading publisher of ad~hoc announcements. Our filings are linked to publicly-traded companies and were published between July~1996 and April~2016. The dataset is split chronologically into a training set~(60\,\%) and a test set~(40\,\%). Announcements were selected such that our corpus includes those that are published in either English (our default analysis) or German, as well as a combined corpus of both. This facilitates later experiments that analyze the sensitivity of news-based forecasts with regard to the underlying language. 

\subsection{Macroeconomic variables}

The ad~hoc announcements are predominantly from German companies. Therefore, we selected a set of macroeconomic indicators with the aim of reflecting the general German economic climate. To account for the share of international firms, we also included a number of European indicators for comparison. \Cref{tbl:indicators} reports our macroeconomic indicators with both monthly and quarterly resolution. 

\begin{table}[htb] 
\centering 
\tiny
\makebox[\textwidth]{
\begin{tabular}{p{3.5cm} ll p{1.3cm} l p{4.5cm} S} 
\toprule
\textbf{Indicator} & \textbf{Region} & \textbf{Abbr.} & \textbf{Resolution (in use)} & \textbf{Source} & \textbf{Description} & \textbf{Obs.} \\ 
\midrule
\textbf{Gross domestic product} & Germany & GDP & Quarterly & OECD & Market value of all final goods and services produced & 86 \\
\textbf{Gross domestic product} & \EU & GDP & Quarterly & ECB & Market value of all final goods and services produced & 86 \\
\textbf{Unemployment rate} & Germany & UR & Monthly & OECD & Number of unemployed people as a percentage of the labor force & 260 \\
\textbf{Consumer price index} & Germany & CPI & Monthly & OECD & Measures prices changes in a basket of goods and services & 260 \\ 
\textbf{Industrial production index} & Germany & IPI & Monthly & OECD & Measures the output in the industrial sector of the economy & 260 \\
\textbf{Business confidence index} & Germany & BCI & Monthly & OECD & Enterprises' assessment of production, orders and stocks & 260 \\ 
\textbf{Consumer confidence index} & Germany & CCI & Monthly & OECD & Households' expectations for the immediate future & 260 \\
\textbf{Federal \num{10} year bond} & Germany & Bund & Monthly & Bloomberg & German government bond denominated in euros & 260 \\
\bottomrule
\end{tabular}%
}%
\caption{List of predicted macroeconomic indicators.} 
\label{tbl:indicators} 
\end{table}

\subsection{Descriptive statistics}


The average length of an ad~hoc announcement over the period amounts to \num{366.15} words. \Cref{tbl:descriptives} lists further descriptives per year. It further indicates an increasing length of the ad~hoc announcements over the research period. The average number of ad~hoc announcements published per year is to 4,072 (for the period 1997--2015).  


\afterpage{
\thispagestyle{empty}
\begin{table}[H] 
\centering 
\tiny
\rotatebox{90}{%
\begin{tabular}{l SS SSSSSS SS} 
\toprule
\textbf{Year} & \textbf{\mcellt{Number\\ disclosures}} & \textbf{\mcellt{Average\\ length}} & \textbf{\mcellt{GDP:\\ Germany}} & \textbf{\mcellt{GDP:\\ \EU}} & \textbf{UR} & \textbf{CPI} & \textbf{IPI} & \textbf{Bund} & \textbf{BCI} & \textbf{CCI} \\ 
\midrule
1996 (Jul--Dec) & 424 & 277.38 & 481.58 & 1498.18 & 9.09 & 1.45 & 77.47  & 6.20  & 98.79 & 98.14 \\ 
1997 & 1524 & 232.96 & 491.77 & 1534.49 & 10.05 & 1.94 & 80.06 & 5.65    & 100.20 & 98.37 \\ 
1998 & 1911 & 281.13 & 504.56 & 1595.85 & 9.41 & 0.91 & 83.23 & 4.53     & 100.52 & 100.34 \\ 
1999 & 3863 & 362.34 & 516.22 & 1669.71 & 8.58 & 0.59 & 84.16 & 4.55     & 99.51  & 100.85 \\ 
2000 & 6954 & 402.03 & 529.12 & 1758.11 & 7.90 & 1.44 & 89.26 & 5.23     & 100.73 & 101.51 \\ 
2001 & 8817 & 296.22 & 544.96 & 1839.15 & 8.01 & 1.98 & 89.64 & 4.83     & 99.21  & 100.56 \\ 
2002 & 7351 & 288.17 & 552.32 & 1902.70 & 8.85 & 1.42 & 88.62 & 4.76     & 98.96  & 99.38 \\ 
2003 & 4676 & 301.33 & 555.02 & 1957.47 & 9.84 & 1.03 & 88.80 & 4.09     & 99.25  & 98.25 \\ 
2004 & 4095 & 309.71 & 567.65 & 2041.11 & 10.01 & 1.67 & 91.56 & 4.06    & 100.32 & 98.65 \\ 
2005 & 4112 & 323.98 & 575.22 & 2114.99 & 11.29 & 1.55 & 94.93 & 3.36    & 100.11 & 99.01 \\ 
2006 & 4221 & 345.60 & 598.31 & 2225.96 & 10.37 & 1.58 & 100.72 & 3.79   & 101.53 & 100.21 \\ 
2007 & 4449 & 376.53 & 628.31 & 2350.20 & 8.80 & 2.30 & 107.59 & 4.24    & 101.94 & 101.74 \\ 
2008 & 4012 & 382.68 & 640.43 & 2408.32 & 7.66 & 2.63 & 107.77 & 3.96    & 100.22 & 99.96 \\ 
2009 & 3500 & 399.41 & 615.07 & 2322.03 & 7.85 & 0.31 & 89.23 & 3.26     & 97.28  & 97.28 \\ 
2010 & 3054 & 432.50 & 645.01 & 2386.20 & 7.07 & 1.10 & 99.33 & 2.74     & 101.09 & 100.60 \\ 
2011 & 2982 & 468.68 & 675.78 & 2449.54 & 5.91 & 2.07 & 107.89 & 2.65    & 101.95 & 101.74 \\ 
2012 & 2903 & 477.53 & 689.57 & 2459.55 & 5.46 & 2.01 & 107.19 & 1.51    & 100.07 & 100.35 \\ 
2013 & 3099 & 454.36 & 706.56 & 2484.55 & 5.31 & 1.50 & 107.32 & 1.62    & 100.34 & 100.44 \\ 
2014 & 3031 & 469.42 & 730.98 & 2531.89 & 5.05 & 0.91 & 109.37 & 1.17    & 100.88 & 101.25 \\ 
2015 & 2811 & 488.65 & 758.20 & 2613.64 & 4.69 & 0.23 & 109.76 & 0.51    & 100.71 & 101.05 \\ 
2016 (Jan--Apr) & 510 & 556.60 & 771.76 & 2645.19 & 4.31 & 0.17 & 111.06 & 0.09   & 100.62 & 100.47 \\ 
\bottomrule
\end{tabular}%
}%
\caption{Descriptive statistics of disclosures and macroeconomic variables. The table lists the average value of each time series by calender year.} 
\label{tbl:descriptives} 
\end{table}
}

\subsection{Semantic constructs}

The proposed projections require a categorization of terms into different semantic structures. The method makes no a~priori assumption regarding the structure and, hence, the final choice can be subject to the intended insights. In fact, the projections support the conventional wordlists from that are widely used in the finance domain or social sciences (the semantic path model merely requires a categorization with neither weights nor mutually-exclusive assignments). Other semantic categories might be beneficial when adapting our approach to other predictive settings, such as marketing. 

In our computational experiments, the constructs should lend themselves to interpretation with regard to perceived sentiments; \eg the valence of the outlook, the uncertainty of the economic development, and the associated confidence by the reporting companies. For this reason, we follow research on financial markets \citep{Nassirtoussi.2014} and draw upon the Loughran-McDonald finance-specific dictionary\footnote{See \url{https://www3.nd.edu/~mcdonald/Word_Lists.html}, last accessed January~12, 2018.}, which categorizes words into different semantic categories that reflect six such perceptions. The \emph{positive} and \emph{negative} categories entail terms that refer to positive and negative assessments; \emph{uncertainty} encodes terms used to signal vagueness or even risks with regard to macroeconomic development; \emph{legal} refers to, \eg, potential liabilities in the future; \emph{strong modal} and \emph{tentative (\ie weak modal)} measure the confidence with regard to statements, conclusions and firm developments. The assignment of words is not unique, as several words belong to more than one semantic category. For instance, all tentative (\ie weak modal) terms are included in the uncertainty list. In total, 132 out of 2,362 entries appear in more than one construct. The choice of this dictionary is beneficial to our setting as the wordlists have been specifically designed to extract qualitative information from regulatory firm disclosures. 

\section{Results} 
\label{sec:results}

This section summarizes the results of our out-of-sample predictive experiments. The predictive performance of the news-based models are evaluated relative to the benchmark models. Here we conducted two different prediction experiments. In the first experiment, the models were trained to predict the raw value of each indicator over multiple prediction horizons. In the second experiment, we looked to test the ability of our news-based approach to predict imminent turning points in the business cycle, where it predicts the change $\Delta{Y}_{i}$ =${Y}_{i+1} - {Y}_{i}$. 

\Cref{tbl:results_quarterly,tbl:results_monthly} compare the prediction performances as a best-of-breed measure. That is, we only report the best-performing time series baseline, the best result from semantic path models, \etc For each best-of-breed model, the performance is measured in terms of root mean squared error~(RMSE). In addition, we validate the robustness of our results by employing the Diebold-Mariano test. The test statistic estimates whether our news-based models obtain statistically significant lower forecast errors compared to the benchmark models. We expect this value frequently to be low due to the small size of the test set and the overall difficulty of macroeconomic predictions. Later, the quality of the forecasts is compared across the different macroeconomic indicators; see \Cref{sec:comparison}. Here we provide the normalized RMSE as it facilitates comparison across different predicted variables.

\begin{table}[!htbp] 
\centering
\footnotesize
\makebox[\textwidth]{
\tiny
\sisetup{input-symbols=(),round-mode=places,round-precision=3}
\begin{tabular}{@{\extracolsep{5pt}} l SSSSSSS} 
\toprule
\textbf{Input/model} & \multicolumn{1}{c}{$\bm{\Delta{Y}}$} & \multicolumn{1}{c}{$\bm{h=1}$} & \multicolumn{1}{c}{$\bm{h=2}$} & \multicolumn{1}{c}{$\bm{h=3}$} & \multicolumn{1}{c}{$\bm{h=4}$} & \multicolumn{1}{c}{$\bm{h=8}$}\\ 
\midrule
\multicolumn{7}{c}{Predicted variable: quarterly Eurozone GDP} \\
\midrule
  Benchmark: lags & 36.37 & 36.37 & 46.365 & 51.112 & 43.654 & 48.217 \\ 
   & {AR6} & {AR6} & {AR6} & {AR6} & {AR1} & {AR1} \\[0.3em] 
  Semantic path model & \bfseries 35.44 & \bfseries 36.108 & 56.506 & 67.177 & 98.954 & 123.697 \\ 
   & \bfseries (0.261) & \bfseries (0.466) & (0.902) & (0.897) & (0.998) & (0.999) \\ 
   & \bfseries {OLS6} & \bfseries {LASSO6} & {LASSO6} & {Ridge6} & {Ridge6} & {Ridge6} \\[0.3em] 
  Semantic features & \bfseries 32.472 & \bfseries 36.108 & 56.506 & 64.948 & 72.079 & 67.379 \\ 
   & \bfseries (0.130) & \bfseries(0.466) & (0.902) & (0.879) & (0.993) & (0.974) \\ 
   & \bfseries {RF6} & \bfseries{LASSO6} & {LASSO6} & {PCR6} & {PCR6} & {PCR6} \\[0.3em] 
  High-dimensional input & 42.319 & 65.529 & 50.119 & \bfseries 48.415 & 64.2 & 82.21 \\
   & (0.846) & (0.995) & (0.681) & \bfseries (0.366) & (0.970) & (0.994) \\ 
   & {LASSO} & {LASSO6} & {LASSO1} & \bfseries {LASSO6} & {LASSO1} & {LASSO6} \\[0.3em] 
  Sensitivity: lemmatization & \bfseries 33.693 & 65.19 & 48.375 & \bfseries 47.968 & 51.121 & 101.873 \\ 
   & \bfseries (0.294) & (0.994) & (0.607) & \bfseries (0.332) & (0.787) & (0.999) \\ 
   & \bfseries {RF} & {LASSO6} & {LASSO1} & \bfseries {LASSO6} & {LASSO1} & {LASSO6} \\[0.3em] 
  Sensitivity: corpus & \bfseries 30.389 & 64.159 & \bfseries 45.767 & \bfseries 45.265 & 58.436 & 67.515 \\ 
   & \bfseries (0.101) & (0.989) & \bfseries (0.472) & \bfseries (0.189) & (0.93) & (0.972) \\ 
   & \bfseries {Complete} & {German} & \bfseries {Complete} & \bfseries {German} & {Complete} & {Complete} \\[0.3em] 
\midrule
\multicolumn{7}{c}{Predicted variable: quarterly Germany GDP} \\
\midrule
  Benchmark: lags & 12.264 & 12.264 & 17.381 & 19.354 & 22.265 & 30.398 \\ 
   & {AR6} & {AR6} & {AR6} & {AR6} & {AR1} & {AR6} \\[0.3em] 
  Semantic path model  & 13.412 & 33.725 & 34.699 & 57.075 & 91.805 & 92.378 \\ 
   & (0.871) & (1.000) & (1.000) & (1.000) & (1.000) & (1.000) \\ 
   & {OLS6} & {LASSO6} & {Ridge6} & {Ridge6} & {Ridge6} & {Ridge6} \\[0.3em] 
  Semantic features & 12.47 & 28.078 & 26.325 & 31.74 & 46.205 & 52.095 \\ 
   & (0.561) & (1.000) & (0.999) & (1.000) & (1.000) & (1.000) \\ 
   & {RF6} & {PCR6} & {PCR6} & {PCR6} & {PCR6} & {PCR6} \\[0.3em] 
  High-dimensional input & \bfseries 9.921 & 23.36 & 26.7 & 44.149 & 59.875 & 63.918 \\ 
   & \bfseries (0.136) & (1.000) & (0.999) & (1.000) & (1.000) & (1.000) \\ 
   & \bfseries {Elastic net} & {LASSO6} & {LASSO6} & {LASSO6} & {LASSO1} & {LASSO6} \\[0.3em] 
  Sensitivity: lemmatization & \bfseries 9.143 & 22.86 & 25.299 & 40.808 & 57.024 & 65.838 \\ 
   & \bfseries (0.092) & (1.000) & (0.997) & (1.000) & (1.000) & (1.000) \\ 
   & \bfseries {LASSO} & {LASSO6} & {LASSO6} & {LASSO6} & {LASSO1} & {LASSO6} \\[0.3em] 
  Sensitivity: corpus & \bfseries 10.637 & 19.681 & 22.766 & 33.858 & 49.914 & 65.936 \\ 
   & \bfseries (0.216) & (0.999) & (0.995) & (1.000) & (1.000) & (1.000) \\ 
   & \bfseries {German} & {Complete} & {German} & {German} & {German} & {German} \\ 
\bottomrule
\end{tabular}
}%
\caption{Comparison of prediction performance across different quarterly macroeconomic indicators. The variable $h$ indicates the number of time steps predicted ahead. Depending on the resolution of the predictive variable, the panels either report long-term forecasts of up to 8~quarters (such that both cases amount to 2~years). The column $\Delta Y$ provides the results for the first difference or change in value from one period ahead to the current period. For each outcome and model type, only the performance of the best-in-breed model is reported (\ie only the best-performing time series model, the best semantic path model, \etc). In each case, the table lists are the root mean squared error. The news-based models that equal or outperform the prediction error of the best benchmark model are in bold for each prediction horizon. In addition, the the $P$-value from the Diebold-Mariano test is stated in the parenthesis when comparing the best-in-breed model to the best baseline, as well as the corresponding model choice including lag structure and dimension reductions (if any).}
\label{tbl:results_quarterly}  
\end{table}

\afterpage{
\thispagestyle{empty}

\centering
\footnotesize
\vspace{-2.5cm}
\makebox[\textwidth]{
\tiny
\sisetup{input-symbols=(),round-mode=places,round-precision=3}
\hspace{-1.5cm}
\begin{tabular}{@{\extracolsep{5pt}} l SSSSSSS S} 
\toprule
\textbf{Input/model} & \multicolumn{1}{c}{$\bm{\Delta{Y}}$} & \multicolumn{1}{c}{$\bm{h=1}$} & \multicolumn{1}{c}{$\bm{h=3}$} & \multicolumn{1}{c}{$\bm{h=6}$} & \multicolumn{1}{c}{$\bm{h=9}$} & \multicolumn{1}{c}{$\bm{h=12}$} & \multicolumn{1}{c}{$\bm{h=24}$}\\ 
\midrule
\multicolumn{7}{c}{Predicted variable: monthly unemployment} \\
\midrule
  Benchmark: lags & 0.07 & 0.07 & 0.17 & 0.359 & 0.661 & 1.105 & 3.419 \\ 
   & {AR1} & {AR1} & {AR1} & {AR1} & {AR1} & {AR1} & {AR1} \\[0.3em]  
  Semantic path model & \bfseries 0.062 & 0.079 & 0.247 & 0.597 & 1.026 & 1.52 & \bfseries 2.777 \\ 
   & \bfseries (0.016) & (0.985) & (1.000) & (1.000) & (1.000) & (1.000) & \bfseries (0.000) \\ 
   & \bfseries {LASSO6} & {OLS1} & {OLS6} & {OLS6} & {OLS6} & {OLS1} & \bfseries {OLS} \\[0.3em]  
  Semantic features & \bfseries 0.062 & 0.081 & 0.277 & 0.668 & 1.07 & 1.531 & \bfseries 2.468 \\ 
   & \bfseries (0.009) & (0.996) & (1.000) & (1.000) & (1.000) & (1.000) & \bfseries (0.000) \\ 
   & \bfseries {LASSO6} & {LASSO1} & {LASSO1} & {LASSO1} & {LASSO1} & {LASSO1} & \bfseries {PCR1} \\[0.3em]  
  High-dimensional input & \bfseries 0.07 & 0.236 & 0.299 & 0.562 & 0.988 & 1.648 & \bfseries 2.3 \\ 
   & \bfseries (0.601) & (1.000) & (1.000) & (1.000) & (1.000) & (1.000) & \bfseries (0.000) \\ 
   & \bfseries {GBM} & {LASSO1} & {LASSO1} & {LASSO1} & {LASSO1} & {PCA-ENET6} & \bfseries {GBM} \\[0.3em] 
  Sensitivity: lemmatization & 0.071 & 0.236 & 0.293 & 0.516 & 0.817 & 1.45 & \bfseries 2.375 \\ 
   & (0.784) & (1.000) & (1.000) & (1.000) & (1.000) & (1.000) & \bfseries (0.000) \\ 
   & {PCR} & {LASSO1} & {LASSO1} & {LASSO1} & {LASSO1} & {PCA-ENET6} & \bfseries {ENET} \\[0.3em] 
  Sensitivity: corpus & \bfseries 0.065 & 0.236 & 0.295 & 0.547 & 0.886 & 1.538 & \bfseries 2.508 \\ 
   & \bfseries (0.083) & (1.000) & (1.000) & (1.000) & (1.000) & (1.000) & \bfseries (0.000) \\ 
   & \bfseries {Complete} & {German} & {German} & {German} & {German} & {German} & \bfseries {Complete} \\[0.3em] 
\midrule
\multicolumn{8}{c}{Predicted variable: monthly inflation rate} \\
\midrule
  Benchmark: lags & 0.29 & 0.29 & 0.497 & 0.656 & 0.711 & 0.793 & 0.749 \\ 
   & {AR1} & {AR1} & {AR1} & {AR6} & {RF6} & {RF6} & {AR6} \\[0.3em] 
  Semantic path model & 0.331 & 0.371 & 0.738 & 0.99 & 1.156 & 1.227 & 0.787 \\ 
   & (0.999) & (1.000) & (1.000) & (1.000) & (1.000) & (1.000) & (0.856) \\ 
   & {OLS6} & {LASSO6} & {LASSO6} & {Ridge6} & {LASSO6} & {OLS6} & {OLS6} \\[0.3em]  
  Semantic features & 0.323 & 0.369 & 0.669 & 0.956 & 1.082 & 1.175 & 0.752 \\ 
   & (0.985) & (1.000) & (1.000) & (1.000) & (1.000) & (1.000) & (0.538) \\ 
   & {PCR} & {RF6} & {RF6} & {RF6} & {PLS-R6} & {PCR1} & {PCR6} \\[0.3em]
  High-dimensional input& \bfseries 0.282 & 0.37 & 0.622 & 0.904 & 0.91 & 0.902 & \bfseries 0.694 \\ 
   & \bfseries (0.175) & (1.000) & (1.000) & (1.000) & (1.000) & (0.998) & \bfseries (0.093) \\ 
   & \bfseries {PCA-ENET6} & {LASSO6} & {LASSO1} & {LSA-PCR} & {LSA-PCR} & {LSA-PCR6} & \bfseries {PCA-GBM1} \\[0.3em]
  Sensitivity: lemmatization & \bfseries 0.288 & 0.382 & 0.623 & 0.907 & 0.909 & 0.896 & \bfseries 0.621 \\ 
   & \bfseries (0.406) & (1.000) & (1.000) & (1.000) & (1.000) & (0.998) & \bfseries (0.005) \\ 
   & \bfseries {PCA-RF} & {LASSO6} & {LASSO1} & {LSA-PCR} & {LSA-PCR6} & {LSA-PCR1} & \bfseries {PCA-GBM6} \\[0.3em]
  Sensitivity: corpus & \bfseries 0.287 & 0.379 & 0.768 & 0.861 & 0.892 & 0.866 & \bfseries 0.631 \\ 
   & \bfseries (0.368) & (1.000) & (1.000) & (1.000) & (1.000) & (0.954) & \bfseries (0.021) \\ 
   & \bfseries {Complete} & {German} & {German} & {German} & {German} & {German} & \bfseries {German} \\ 
\midrule
\multicolumn{8}{c}{Predicted variable: monthly industrial production rate} \\
\midrule
  Benchmark: lags & 1.827 & 1.827 & 3.505 & 5.703 & 6.532 & 6.922 & 9.816 \\ 
   & {AR1} & {AR1} & {AR6} & {AR6} & {AR6} & {AR1} & {AR6} \\[0.3em]
  Semantic path model & 1.905 & 1.851 & \bfseries 3.497 & \bfseries 5.44 & \bfseries 6.12 & 7.063 & 12.631 \\ 
   & (0.947) & (0.739) & \bfseries (0.439) & \bfseries (0.024) & \bfseries (0.169) & (0.663) & (1.000) \\ 
   & {Ridge1} & {LASSO1} & \bfseries {LASSO6} & \bfseries {OLS6} & \bfseries {ENET6} & {Ridge6} & {Ridge} \\[0.3em]
  Semantic features & 1.884 & 1.851 & \bfseries 3.497 & \bfseries 5.468 & \bfseries 6.12 & \bfseries 6.896 & 10.987 \\ 
   & (0.879) & (0.739) & \bfseries (0.439) & \bfseries (0.024) & \bfseries (0.169) & \bfseries (0.464) & (0.999) \\ 
   & {PCR} & {LASSO1} & \bfseries {LASSO6} & \bfseries {LASSO6} & \bfseries {ENET6} & \bfseries {PLS-R6} & {PCR1} \\[0.3em]
  High-dimensional input& \bfseries 1.812 & 2.045 & 3.521 & \bfseries 5.392 & \bfseries 5.521 & \bfseries 5.064 & 10.37 \\ 
   & \bfseries (0.128) & (0.984) & (0.546) & \bfseries (0.268) & \bfseries (0.020) & \bfseries (0.000) & (0.946) \\ 
   & \bfseries {LSA-LASSO6} & {LASSO1} & {LASSO1} & \bfseries {PCA-LASSO6} & \bfseries {PCA-GBM} & \bfseries {PLS-R1} & {Ridge1} \\[0.3em]
  Sensitivity: lemmatization & \bfseries 1.817 & 2.035 & 3.522 & \bfseries 5.364 & \bfseries 5.403 & \bfseries 4.978 & 10.415 \\ 
   & \bfseries (0.329) & (0.983) & (0.548) & \bfseries (0.234) & \bfseries (0.006) & \bfseries (0.000) & (0.978) \\ 
   & \bfseries {LSA-GBM} & {LASSO1} & {LASSO1} & \bfseries {PCA-LASSO6} & \bfseries {ENET1} & \bfseries {PLS-R1} & {PCR6} \\[0.3em]
  Sensitivity: corpus & \bfseries 1.814 & 1.958 & \bfseries 3.435 & \bfseries 5.336 & \bfseries 5.238 & \bfseries 4.254 & \bfseries 5.557 \\ 
   & \bfseries (0.107) & (0.928) & \bfseries (0.323) & \bfseries (0.125) & \bfseries (0.003) & \bfseries (0.000) & \bfseries (0.000) \\ 
   & \bfseries {German} & {German} & \bfseries {Complete} & \bfseries {Complete} & \bfseries {Complete} & \bfseries {German} & \bfseries {German} \\ 
\end{tabular}
}

\vspace{0.3cm}
\emph{continued on next page}
\newpage

\begin{table}[H] 
\centering
\footnotesize
\vspace{-3cm}
\hspace{-1.5cm}
\makebox[\textwidth]{
\tiny
\sisetup{input-symbols=(),round-mode=places,round-precision=3}
\begin{tabular}{@{\extracolsep{5pt}} l SSSSSSS S}
\midrule
\multicolumn{8}{c}{Predicted variable: monthly business confidence index} \\
\midrule
  Benchmark: lags & 0.057 & 0.057 & 0.318 & 0.791 & 0.96 & 0.955 & 1.023 \\ 
   & {AR6} & {AR6} & {AR6} & {AR6} & {AR6} & {AR6} & {AR1} \\[0.3em]
  Semantic path model & 0.061 & \bfseries 0.057 & 0.334 & 0.799 & \bfseries 0.878 & \bfseries 0.847 & 1.146 \\ 
   & (0.943) & \bfseries (0.543) & (0.85) & (0.584) & \bfseries (0.020) & \bfseries (0.006) & (1.000) \\ 
   & {OLS6} & \bfseries {LASSO6} & {OLS6} & {OLS6} & \bfseries {LASSO6} & \bfseries {Ridge6} & {OLS6} \\[0.3em]
  Semantic features & 0.063 & \bfseries 0.057 & 0.334 & 0.812 & \bfseries 0.878 & \bfseries 0.847 & 1.515 \\ 
   & (0.989) & \bfseries (0.543) & (0.858) & (0.71) & \bfseries (0.020) & \bfseries (0.006) & (1.000) \\ 
   & {LASSO6} & \bfseries {LASSO6} & {LASSO6} & {LASSO6} & \bfseries {LASSO6} & \bfseries {Ridge6} & {{USM1}} \\[0.3em]
  High-dimensional input& 0.218 & 0.28 & 0.784 & 1.291 & 1.199 & 0.976 & 1.21 \\ 
   & (1.000) & (1.000) & (1.000) & (1.000) & (0.986) & (0.586) & (0.987) \\ 
   & {LASSO6} & {ENET1} & {ENET1} & {LASSO6} & {PCA-RF6} & {PCA-RF6} & {PCA-PCR} \\[0.3em]
  Sensitivity: lemmatization & 0.235 & 0.28 & 0.784 & 1.323 & 1.286 & 1.033 & 1.213 \\ 
   & (1.000) & (1.000) & (1.000) & (1.000) & (0.997) & (0.811) & (0.988) \\ 
   & {GBM6} & {ENET1} & {ENET1} & {LASSO6} & {PCA-RF6} & {PCA-GBM6} & {PCA-PCR} \\[0.3em]
  Sensitivity: corpus & 0.232 & 0.281 & 0.795 & 1.272 & 1.115 & 0.996 & 1.09 \\ 
   & (1.000) & (1.000) & (1.000) & (1.000) & (0.971) & (0.681) & (0.81) \\ 
   & {German} & {German} & {Complete} & {Complete} & {Complete} & {German} & {Complete} \\ 
\midrule
\multicolumn{8}{c}{Predicted variable: monthly consumer confidence index} \\
\midrule
  Benchmark: lags & 0.085 & 0.085 & 0.454 & 0.998 & 1.212 & 1.206 & 1.396 \\ 
   & {AR6} & {AR6} & {AR6} & {AR6} & {AR6} & {AR6} & {AR1} \\[0.3em]
  Semantic path model & 0.093 & 0.089 & 0.516 & 1.136 & 1.263 & \bfseries 1.074 & 1.841 \\ 
   & (0.996) & (0.824) & (0.978) & (0.984) & (0.796) & \bfseries (0.004) & (1.000) \\ 
   & {OLS6} & {LASSO6} & {LASSO6} & {LASSO6} & {OLS6} & \bfseries {OLS6} & {OLS6} \\[0.3em]
  Semantic features & 0.094 & 0.089 & 0.516 & 1.136 & 1.226 & \bfseries 1.061 & 2.003 \\ 
   & (0.999) & (0.824) & (0.978) & (0.984) & (0.566) & \bfseries (0.002) & (1.000) \\ 
   & {LASSO6} & {LASSO6} & {LASSO6} & {LASSO6} & {USM6} & \bfseries {USM6} & {RF6} \\[0.3em]
  High-dimensional input& 0.245 & 0.293 & 0.79 & 1.217 & 1.294 & 1.229 & \bfseries 1.375 \\ 
   & (1.000) & (1.000) & (1.000) & (0.996) & (0.946) & (0.612) & \bfseries (0.385) \\ 
   & {RF6} & {LASSO1} & {LASSO1} & {RF1} & {PCA-Ridge1} & {PCA-GBM1} & \bfseries {LSA-LASSO6} \\[0.3em]
  Sensitivity: lemmatization & 0.246 & 0.293 & 0.79 & 1.235 & 1.295 & 1.26 & \bfseries 1.389 \\ 
   & (1.000) & (1.000) & (1.000) & (0.998 & (0.916) & (0.767) & \bfseries (0.465) \\ 
   & {ENET6} & {LASSO1} & {LASSO1} & {LASSO1} & {PCA-ENET1} & {PCA-LASSO1} & \bfseries {LSA-PCR6} \\[0.3em]
  Sensitivity: corpus & 0.245 & 0.294 & 0.736 & 1.188 & \bfseries 1.141 & 1.25 & \bfseries 1.371 \\ 
   & (1.000) & (1.000) & (1.000) & (0.987) & \bfseries (0.054) & (0.735) & \bfseries (0.359) \\ 
   & {Complete} & {German} & {German} & {German} & \bfseries {Complete} & {Complete} & \bfseries {Complete} \\
\midrule
\multicolumn{8}{c}{Predicted variable: federal 10-year bond} \\
\midrule
  Benchmark: lags & 0.23 & 0.266 & 0.62 & 1.076 & 1.399 & 1.723 & 2.207 \\ 
   & {RF6} & {AR1} & {AR6} & {AR1} & {AR1} & {AR6} & {AR6} \\[0.3em] 
  Semantic path model & \bfseries 0.226 & 0.444 & 1.042 & 1.866 & 2.092 & 2.427 & 2.484 \\ 
   & \bfseries (0.303) & (1.000) & (1.000) & (1.000) & (1.000) & (1.000) & (1.000) \\ 
   & \bfseries {LASSO1} & {LASSO1} & {OLS1} & {OLS1} & {OLS1} & {Ridge1} & {OLS} \\[0.3em] 
  Semantic features & \bfseries 0.226 & 0.392 & 1.048 & 1.517 & 2.068 & 2.197 & 2.43 \\ 
   & \bfseries (0.303) & (1.000) & (1.000) & (1.000) & (1.000) & (1.000) & (1.000) \\ 
   & \bfseries {LASSO1} & {PLS-R6} & {LASSO1} & {PCR6} & {PLS-R1} & {PCR1} & {PLS-R} \\[0.3em] 
  High-dimensional input& \bfseries 0.209 & 0.434 & 1.184 & 1.942 & 2.147 & 2.144 & \bfseries 2.121 \\ 
   & \bfseries (0.004) & (1.000) & (1.000) & (1.000) & (1.000) & (1.000) & \bfseries (0.001) \\ 
   & \bfseries {LSA-LASSO} & {LASSO1} & {LASSO1} & {LASSO1} & {LASSO1} & {ENET1} & \bfseries {PCA-ENET} \\[0.3em] 
  Sensitivity: lemmatization & 0.212 & 0.441 & 1.158 & 2.035 & 2.262 & 2.046 & \bfseries 2.117 \\ 
   & \bfseries (0.008) & (1.000) & (1.000) & (1.000) & (1.000) & (1.000) & \bfseries (0.000) \\ 
   & \bfseries {LSA-Ridge} & {LASSO1} & {LASSO1} & {LASSO1} & {PLS-R6} & {ENET1} & \bfseries {PCA-ENET} \\[0.3em] 
  Sensitivity: corpus & 0.211 & 0.422 & 1.07 & 1.771 & 1.886 & 1.868 & \bfseries 1.788 \\ 
   & \bfseries (0.004) & (1.000) & (1.000) & (1.000) & (1.000) & (1.000) & \bfseries (0.000) \\ 
   & \bfseries {Complete} & {German} & {German} & {German} & {German} & {German} & \bfseries {Complete} \\ 
\bottomrule
\end{tabular}
}%
\caption{Comparison of prediction performance across different monthly macroeconomic indicators. The variable $h$ indicates the number of time steps predicted ahead. Depending on the resolution of the predictive variable, the panels either report long-term forecasts of up to 24~months (such that both cases amount to 2~years). The column $\Delta Y$ provides the results for the first difference or change in value from one period ahead to the current period. For each outcome and model type, only the performance of the best-in-breed model is reported (\ie only the best-performing time series model, the best semantic path model, \etc). In each case, the table lists are the root mean squared error. The news-based models that equal or outperform the prediction error of the best benchmark model are in bold for each prediction horizon. In addition, the the $P$-value from the Diebold-Mariano test is stated in the parenthesis when comparing the best-in-breed model to the best baseline, as well as the corresponding model choice including lag structure and dimension reductions (if any).}
\label{tbl:results_monthly}  
\end{table}
}

\subsection{Prediction performance across macroeconomic indicators}

\subsubsection{Eurozone GDP}

\Cref{tbl:results_quarterly} lists the RMSE values obtained from the predictive experiments undertaken for the Eurozone GDP. Several news-based models are able to surpass the accuracy of the benchmarks for a variety of the prediction horizons. First, in the case of the first-difference and \num{1} quarter ahead prediction horizon, both the semantic path model and semantic features prove to be more accurate than both the high-dimensional input and benchmark models. Relative to best high-dimensional input, the RMSE of \num{42.319} and \num{65.529} represents 23\,\% and 44.8\,\% reduction, respectively. 

When looking at the long-term predictions, the LASSO-trained high-dimensional input model outperforms the benchmark for the \num{3} quarters ahead prediction horizon, in this case achieving a RMSE of \num{48.415} which is a 5\,\% reduction relative to the baseline. Furthermore, the high-dimensional input prove to be more accurate over the medium and long-term prediction horizons than the semantic paths models. 

\subsubsection{German GDP}

The prediction errors for the German GDP are as follows. We find the elastic net trained on the high-dimensional input the most accurate for first-difference prediction horizon, achieving a RMSE of \num{9.921}. We observe improvements over the benchmark only fore aforementioned horizons. This is especially interesting, since it differs from the prediction results for the whole Eurozone, where we achieved consistent improvements.

\subsubsection{Unemployment}

In terms of unemployment rate, the news-based models outperform the benchmark over the first-difference prediction. We find both the semantic path model, as well as the semantic features, to be the most accurate with a RMSE of \num{0.062}. Furthermore, the DM test results are highly significant for both models derived from semantic inputs.

We observe that news-based models cannot beat the benchmarks for medium-term predictions, while we again gain improvements from using news in the long-term. Here the news-based models consistently outperform the baseline for the \num{24} months ahead prediction horizon. The highest predictive accuracy stems from gradient boosting applied to high-dimensional input with a prediction error of \num{2.300}. The best of the benchmarks reported only an RMSE of \num{3.419}; this results in a forecast error reduction of \SI{32.7}{\percent}. Furthermore, the Diebold-Mariano test results are highly statistically significant for the \num{24} months prediction horizon for all news-based models. The semantic path model attains only an improvement of \SI{18.8}{\percent}, yet while being fully interpretable.


\subsubsection{Inflation}

We find support for the fact that the news-based models are superior to the benchmarks for the first difference prediction horizons. The high-dimensional input accounts for the lowest RMSE (\num{0.209}). Furthermore, the Diebold-Mariano test is statistically significant. For the prediction horizons \num{1} to \num{12} months ahead, our news-based models attain similar levels of accuracy however are unable to best the leading benchmark. Further improvements in predictive performance are evident for the \num{24} months ahead prediction horizon. Here the high-dimensional input with PCA dimensionality reduction yields the lowest RMSE of \num{0.694}, \ie an improvement of \SI{7.3}{\percent} over the time series baseline. The observed results indicate that the financial news stemming from listed corporations can be utilized to capture the rate of change in consumer prices in the short-term, and that these, continue to be predictive over the long-term.
  
\subsubsection{Industrial production}

The news-based models outperform the benchmarks for a majority of tested prediction horizons. In particular, the news-based model achieve a lower forecast error for the mid-range prediction horizons. The experiments for the first-difference prediction reveal that the high-dimensional input with LSA preprocessing obtains a RMSE of \num{1.812} (\ie an improvement of \SI{0.8}{\percent}. The news-based models outperform the benchmark for the 3--12 months ahead prediction horizons. Both the semantic path model and the semantic features attain the best RMSE (\num{3.497}), for the \num{3} months ahead prediction. The high-dimensional input is the most accurate for the 6--12 months ahead prediction horizon. The $p$-values from the individual DM tests indicate that predicted values from our news-based system are significantly more accurate over the before-mentioned prediction horizons. 


\subsubsection{Business confidence}

The semantic path model, as well as semantic features, are able to outperform the benchmark and the high-dimensional input across a number of prediction horizons. Both methods register a the lowest RMSE for the \num{1} month (\num{0.057}), \num{9} month (\num{0.878}) and \num{12} month (\num{0.847}) ahead prediction horizons. The latter amounts to an improvement of \SI{11.3}{\percent}. Furthermore, these model significantly out-perform the high-dimensional input across all predictions horizons. The business confidence index is calculated by conducting surveys of business leaders who know the current state of the corporate sector. The semantic path model can adapt at capturing the sentiment of the individual announcements; \ie the same sentiment would be reflected in the index. The lack of long-term predictive power for the high-dimensional inputs could be due to the fact that the news content might already be factored into their survey answers, therefore reducing potential lag affects. Overall, the semantic models seems better suited for predictions of the business confidence index than the high-dimensional inputs.

\subsubsection{Consumer confidence}

We investigate the prognostic capability of financial news for the consumer confidence index. Overall, the predictive performance of the news-based models does not diverge greatly from those of the benchmarks. Of the news-based models, both semantic models prove to be more accurate than the high-dimensional input. The news-based models do narrowly outperform the benchmark for prediction horizons greater than \num{9} months ahead.

\subsubsection{Government bonds}


The news-based models perform best relative to the benchmarks for the first-difference prediction horizons. For instance, the high-dimensional input attains the lowest overall RMSE (\num{0.209}, \ie a minus of \SI{9.1}{\percent} over the best baseline), closely followed by both the semantic models (\num{0.226}, \ie a minus of \SI{1.7}{\percent}), while the RMSE of the best benchmark is \num{0.230}. The high-dimensional input is also able to outperform best the benchmark for the \num{24} months ahead prediction horizon. The Diebold-Mariano test results for the high-dimensional input are all highly statistically significant. 


\subsection{Interpretation of latent semantic structures}

\Cref{fig:interpretation} presents the results from the semantic path model and decomposes the forecasts into different constructs in order to facilitate interpretation. That is, we compute the value of each construct -- given by predicted value $\psi_i z_i$ -- and compare these figures across the different semantic categories according to the Loughran-McDonald finance-specific dictionary. These categories are: (i)~positivity/negativity of economic outlook, (ii)~uncertainty regarding economic development, (iii)~legal risk for operations and (iv)~confidence/modality of the disclosing company.

Altogether, we find the positive construct has the strongest impact on both Eurozone and German GDP and, to a lesser extent, on consumer confidence, industrial production and government bonds. The negative construct, on the other hand, affects the unemployment rate, inflation, industrial production, consumer confidence and government bonds. A rise in negative news could see firms downsize operations and cut investment spending, resulting in lower levels of employment. Reduced confidence in the economic outlook could see consumers also cut spending placing downward pressure on prices. 

\afterpage{
\thispagestyle{empty}
\begin{figure}[H]
\centering
\footnotesize
\makebox[1.3\textwidth]{
\hspace{-2cm}
\begin{tabular}{cc}
(a)~Eurozone GDP & (b)~German GDP \\
\includegraphics[width=.45\textwidth,trim={0 0 4cm 0},clip]{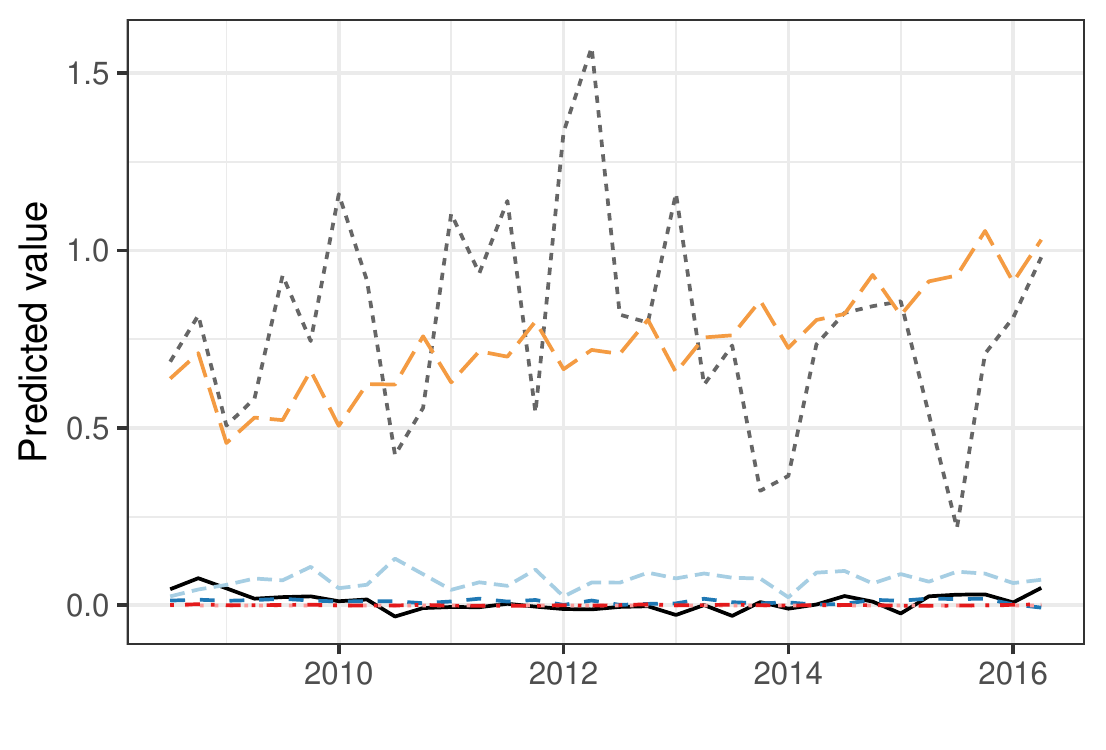} & \includegraphics[width=.65\textwidth]{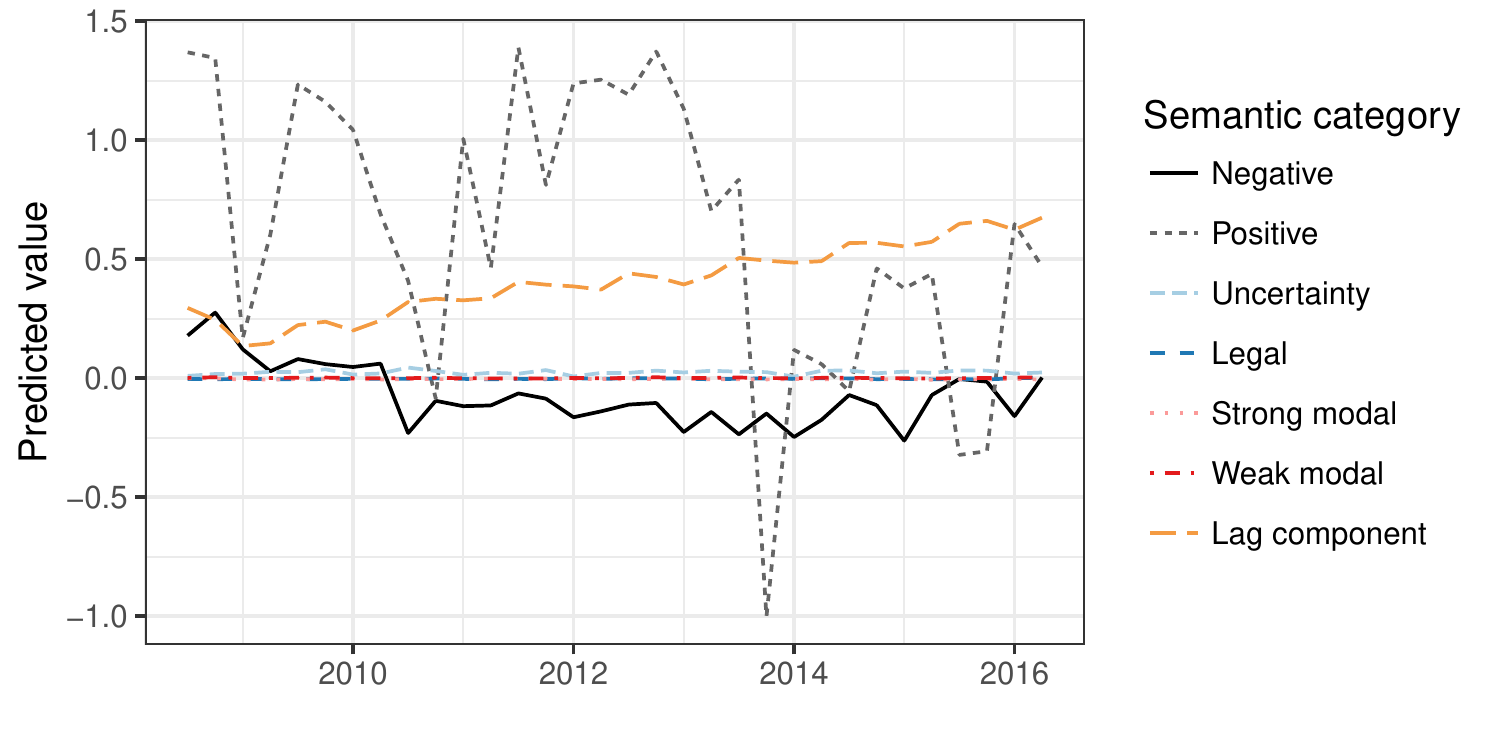} \\
(c)~Unemployment rate & (d)~Inflation (consumer price index) \\
\includegraphics[width=.45\textwidth,trim={0 0 4cm 0},clip]{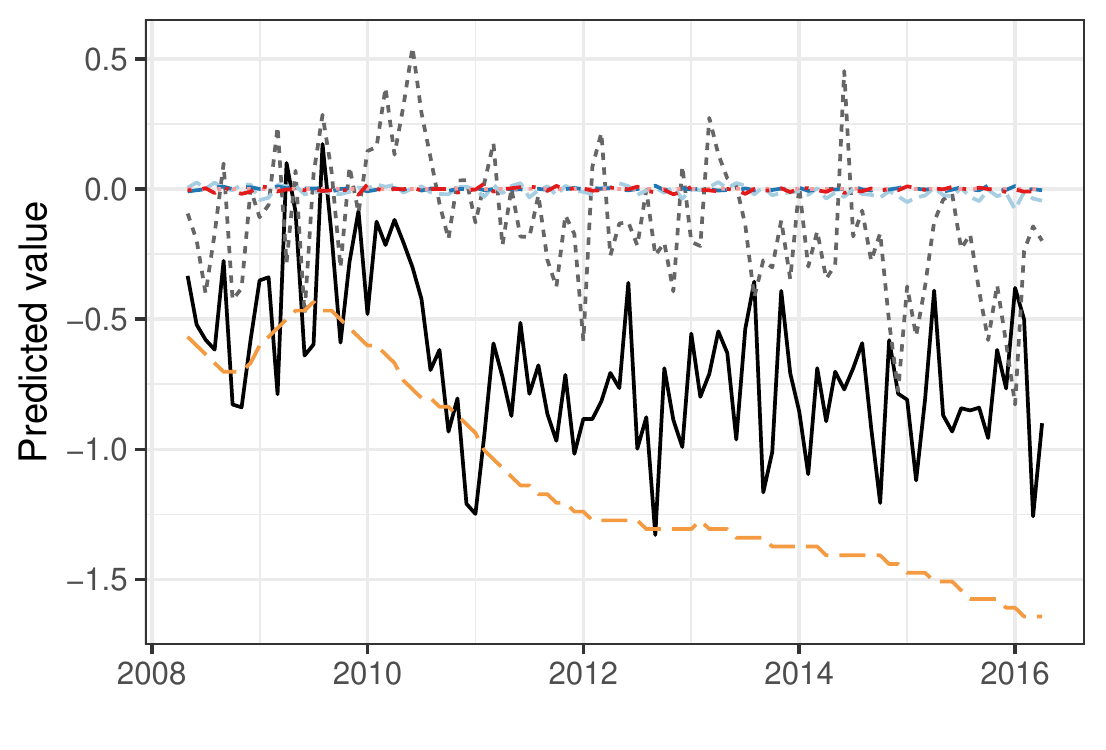} & \includegraphics[width=.45\textwidth,trim={0 0 4cm 0},clip]{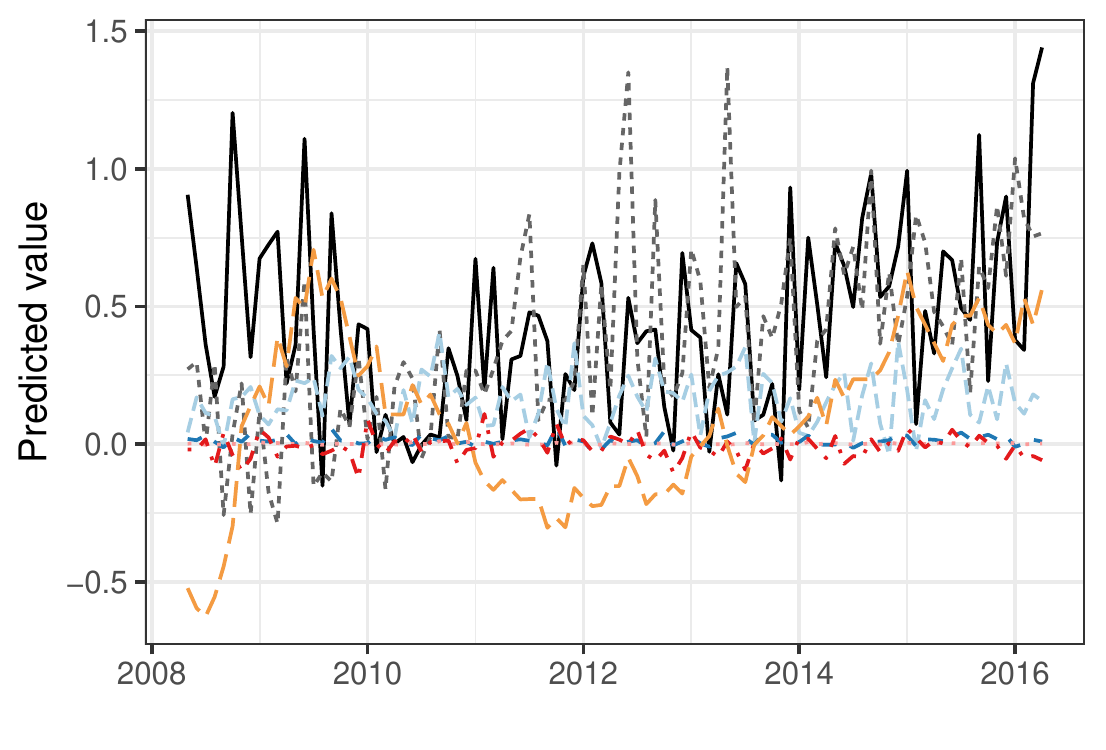} \\
(e)~Industrial production index & (f)~Business confidence index \\
\includegraphics[width=.45\textwidth,trim={0 0 4cm 0},clip]{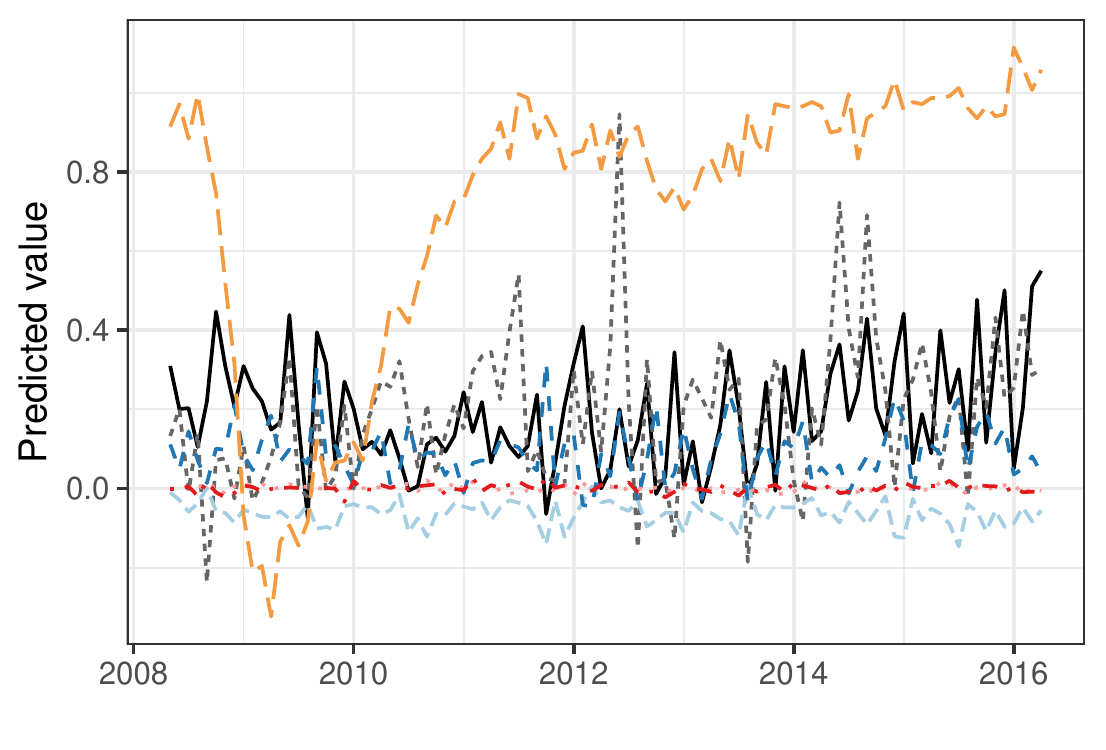} & \includegraphics[width=.45\textwidth,trim={0 0 4cm 0},clip]{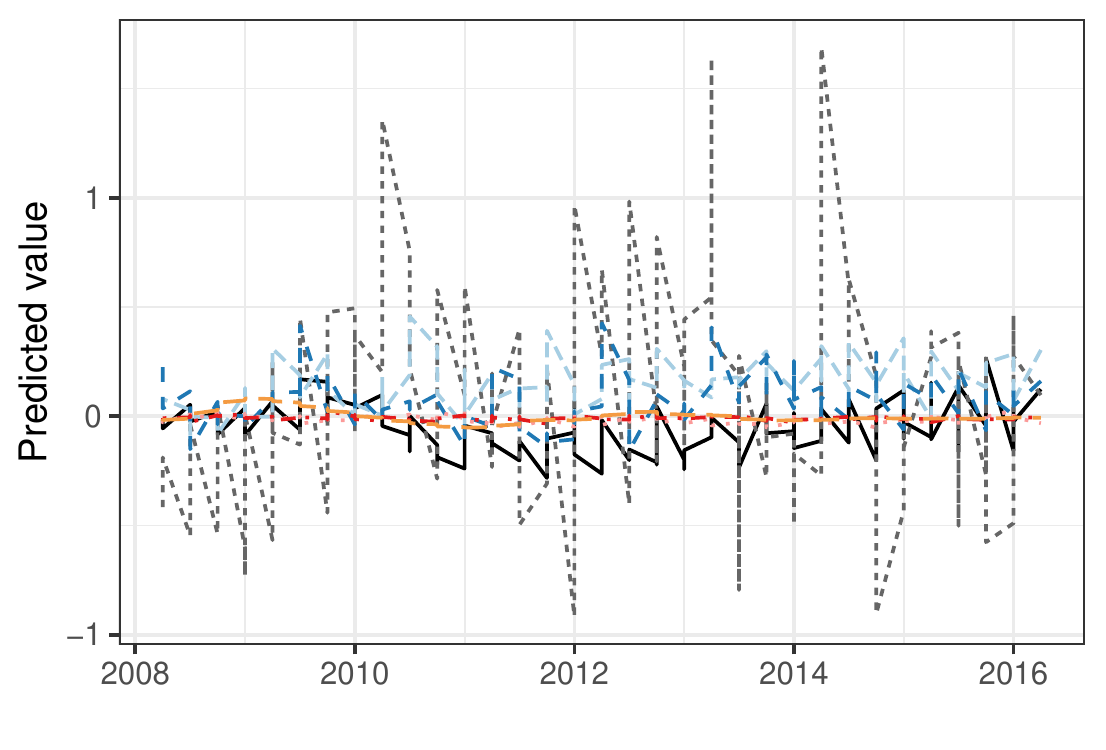} \\
(g)~Consumer confidence index & (h)~Government bond \\
\includegraphics[width=.45\textwidth,trim={0 0 4cm 0},clip]{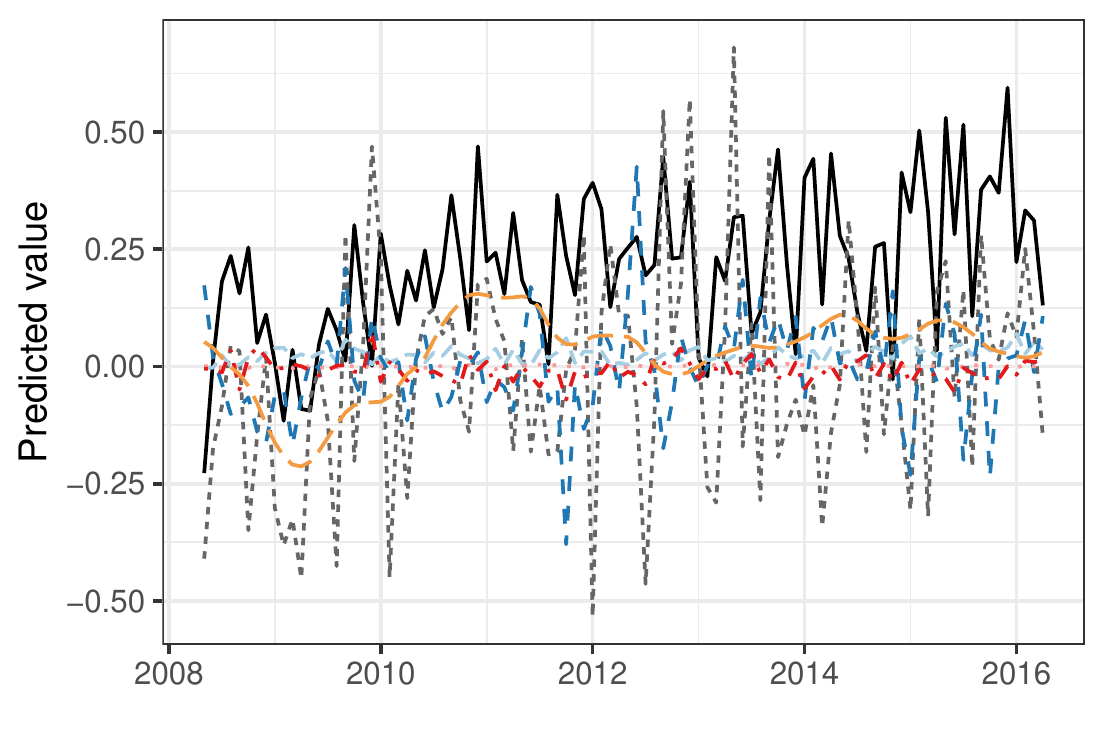} & \includegraphics[width=.45\textwidth,trim={0 0 4cm 0},clip]{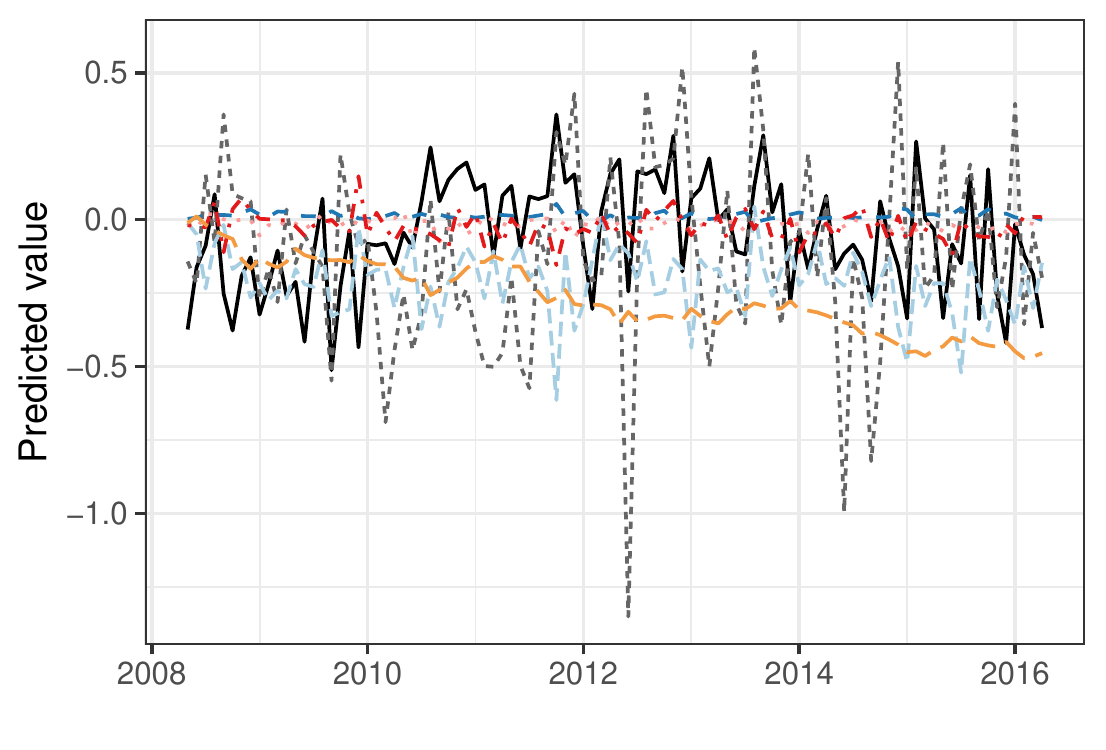} 
\end{tabular}
}
\caption{Illustrated interpretability of semantic path models. The value of each construct -- given by predicted value $\psi_i z_i$ -- is plotted over the course of the test set. Note that the inner model from our semantic path model operates on normalized constructs $z_i$, which is even beneficial for the above comparisons. The latent structures directly lend to interpretation with respect to our semantic dimension. For reasons of comparability, we apply the same estimation technique to all macroeconomic indicators: the inner model is estimated via ordinary least squares for a prediction 1-year ahead based on the English corpus with 1 lag.}
\label{fig:interpretation}
\end{figure}
}

The lag component is strongly predictive for all variables except business confidence. Potentially, a change in perceptions at a corporate level would already be factored into the index by business leaders. The remaining indicators would be slower to adjust. We find the uncertainty construct to be predictive for inflation, the confidence indices and government bonds. Consumers, firms and financial markets all have an aversion to uncertainty. The legal construct seems to be predictive for industrial production, business and consumer confidence. A possibility explanation is that firms facing legal difficulties would reduce output and other business investment impacting both the suppliers and consumers that function within the economic stream. Finally, we find the modal constructs to be a weak predictor across all of the chosen indicators.

\subsection{Sensitivity analysis}

\Cref{tbl:results_quarterly,tbl:results_monthly} entail additional sensitivity checks. First, we compare stemming as a default in contrast to lemmatization. While the latter adapts correctly to the semantic context, it suffers from larger predictor matrices and potentially higher risk of overfitting. However, our results suggest no clear indication concerning which approach is preferable with regard to our task. Second, we performed all analyses based on the English corpus and now compare this choice to both a corpus with German articles and a combined one. Again, we find no clear evidence that one consistently outperforms the other. A potential reason  for this parity stems from the fact that the majority of firms publish in both languages and any potential differences therefore are relatively minor.

\subsection{Comparison}
\label{sec:comparison}

While the previous results compared the use of time series models to a machine learning approach, we now turn our attention to the overall capacity of foreseen macroeconomic development. For this reason, we computed the normalized RMSE for all macroeconomic indicators as it facilitates comparisons between different outcome variables and their respective scales. Formally, it is defined by $\mathrm{NRMSE} = \cfrac{\mathrm{RMSE}}{Y_{\max} - Y_{\min}}$.

The normalized RMSE results are detailed in \Cref{tbl:nrmse}. We find the Eurozone GDP results to have lower prediction errors than German GDP. Potential reasons could be the export orientation of German companies, or the higher volatility of the Eurozone GDP, which can be better explained by production-related information from news disclosures. The prediction accuracy for the business confidence index relative to consumer confidence is also better. Given the direct link between the business leaders who are surveyed to estimate the index and the source of the ad~hoc announcements, a higher rate of accuracy for the business confidence index could be expected.

In comparing the performance of both the semantic models relative to the high-dimensional input, we find the results varying depending on the indicator and prediction horizon. The semantic models consistently attain lower prediction errors for the confidence indicies across all the prediction horizons except for the very long run. {In fact, the semantic path model accomplishes the best forecast in 6 prediction experiments, while the semantic features do so in 8 cases and the high-dimensional in 12, yet the latter with hardly any interpretability. This is interesting in the light that our semantic path model enforces a linear structure among constructs and thus excludes potentially non-linear relationships as used by predictive models with semantic features or the universal structure model. While non-linearity introduces additional flexibility, our setting is extremely overspecified and the smaller parameter space of linear models can often be beneficial.}  Moreover, we see that traditional approaches to unsupervised dimensionality reduction rarely yield the best-of-breed model; conversely, our domain knowledge can outperform them frequently.

 
The overall performance of the news-based models varies depending on the macroeconomic variable and prediction horizon. Over the short- and medium term, the benchmark models tend to be superior in terms of consumer-centered variables, such as unemployment rate or inflation. However, financial news reflects both the current health and future expectations of corporations. As a consequence, news-based models appear especially useful for product-related indicators, such as the actual industrial product rate and confidence indices. Finally, in terms of long-term predictions, the news-based models tend to outperform the benchmark models. Altogether, this provides ample evidence of the strength of the predictive properties of our news-based inputs. 

\begin{table}[!htbp] 
\centering
\tiny
\makebox[\textwidth]{
\footnotesize
\sisetup{input-symbols=(),round-mode=places,round-precision=1}
\begin{tabular}{@{\extracolsep{5pt}} ll SSSSSSS} 
\toprule
\textbf{Predicted variable} & \textbf{Model} & \multicolumn{7}{c}{\textbf{Normalized RMSE}}\\ 
\midrule
&& \multicolumn{7}{c}{\textbf{Quarterly resolution}} \\
\cmidrule(lr){2-9}
& \multicolumn{1}{c}{$\bm{\Delta{Y}}$} & \multicolumn{1}{c}{$\bm{h=1}$} & \multicolumn{1}{c}{$\bm{h=2}$} & \multicolumn{1}{c}{$\bm{h=3}$} & \multicolumn{1}{c}{$\bm{h=4}$} & \multicolumn{1}{c}{$\bm{h=8}$} \\
\midrule
Eurozone GDP 
 & Semantic path model & 12.0 &  8.0 & 12.5 & 16.2 & 23.9 & 37.9 \\ 
 & Semantic features &  11.0 &  8.0 & 12.5 & 15.7 & 17.4 & 20.6 \\ 
 & High-dimensional input & 14.3 & 14.5 & 11.1 &  11.7 & 15.5 & 25.2 \\[0.3em] 
German GDP 
 & Semantic path model & 21.1 & 18.5 & 19.0 & 32.0 & 57.3 & 77.1 \\ 
 & Semantic features & 19.6 & 15.4 & 14.4 & 17.8 & 28.8 & 43.5 \\ 
 & High-dimensional input &  15.6 & 12.8 & 14.6 & 24.7 & 37.4 & 53.4 \\
\midrule
&& \multicolumn{7}{c}{\textbf{Monthly resolution}} \\
\cmidrule(lr){2-9}
& \multicolumn{1}{c}{$\bm{\Delta{Y}}$} & \multicolumn{1}{c}{$\bm{h=1}$} & \multicolumn{1}{c}{$\bm{h=3}$} & \multicolumn{1}{c}{$\bm{h=6}$} & \multicolumn{1}{c}{$\bm{h=9}$} & \multicolumn{1}{c}{$\bm{h=12}$} & \multicolumn{1}{c}{$\bm{h=24}$}\\
\midrule 
Unemployment 
   & Semantic path model &  15.4 & 2.2 & 6.7 & 16.1 & 27.7 & 41.1 & 99.2 \\ 
   & Semantic features &  15.4 & 2.2 & 7.5 & 18.1 & 28.9 & 41.4 & 88.1 \\ 
   & High-dimensional input & 17.6 & 6.5 & 8.1 & 15.2 & 26.7 & 44.5 &  82.1 \\[0.3em] 
Inflation 
   & Semantic path model & 18.9 & 9.7 & 20.4 & 34.1 & 39.9 & 42.3 & 29.3 \\ 
   & Semantic features & 18.5 & 9.6 & 18.5 & 33 & 37.3 & 40.5 & 28.0 \\ 
   & High-dimensional input &  16.1 & 9.7 & 17.2 & 31.2 & 31.4 & 31.1 &  25.9 \\[0.3em]%
Industrial production 
   & Semantic path model & 16.6 & 6.6 &  12.4 & 19.4 & 21.8 & 28.7 & 92.9 \\ 
   & Semantic features & 16.4 & 6.6 &  12.4 & 19.5 & 21.8 & 28.0 & 80.8 \\ 
   & High-dimensional input &  15.8 & 7.3 & 12.5 &  19.2 &  19.6 &  20.6 & 76.2 \\[0.3em]
Business confidence 
   & Semantic path model & 4.6 &  0.9 & 5.1 & 12.3 &  13.5 &  13.7 & 32.6 \\ 
   & Semantic features & 4.7 &  0.9 & 5.2 & 12.5 &  13.5 &  13.7 & 44.9 \\ 
   & High-dimensional input & 16.4 & 4.3 & 12.1 & 19.9 & 18.5 & 15.7 & 34.4 \\[0.3em] 
Consumer confidence 
   & Semantic path model & 6.6 & 1.4 & 8.1 & 17.8 & 19.8 & 17.3 & 61.7 \\ 
   & Semantic features & 6.6 & 1.4 & 8.1 & 17.8 & 19.9 &  17.2 & 67.1 \\ 
   & High-dimensional input & 17.3 & 4.6 & 12.4 & 19.0 & 20.2 & 19.8 &  46.1 \\[0.3em] 
Government bonds 
   & Semantic path model & 21.5 & 9.8 & 24.2 & 50.2 & 56.2 & 65.3 & 71.3 \\ 
   & Semantic features & 21.5 & 8.7 & 24.3 & 40.8 & 55.6 & 59.1 & 69.7 \\ 
   & High-dimensional input &  19.8 & 9.6 & 27.5 & 52.2 & 57.7 & 57.7 &  60.9 \\ 
\bottomrule
\end{tabular}
}
\caption{Summary of text-based prediction performance across different macroeconomic indicators. Here the normalized RMSE is reported in order to facilitate comparison across outcomes (and their respective scales). The variable $h$ indicates the number of time steps predicted ahead. Depending on the resolution of the predictive variable, the panels either report long-term forecasts of up to 8~quarters or 24~months (such that both cases amount to 2~years). The column $\Delta Y$ provides the results for the first difference or change in value from one period ahead to the current period. For each outcome and model type, only the performance of the best-in-breed model is reported (\ie only the best-performing time series model, the best semantic path model, \etc). 
}
\label{tbl:nrmse}  
\end{table}

\section{Managerial implications} 
\label{sec:discussion}


The economic climate has long-ranging effects on firm operations. As a consequence, managers and policymakers alike are required to form expectations, hedge potential operational risks from economic downturns and adapt their operations accordingly. Our method thus assists managers in making such predictions and provides a cost-effective alternative to expert forecasts. Firm managers can utilize our predictions in order to, among other things, set long-term strategic plans, decide on investment levels, and manage human resources. Beyond that, managers in public sector institutions may benefit from our long-term predictions when setting fiscal policy, such as government spending, borrowing and taxation, and monetary policy concerning to control of the supply of money within the economy. Our innovative, news-based methodology also provides investment managers a greater ability to anticipate the impact of short-term economic fluctuations on their portfolios and, if necessary, take the necessary precautions to hedge against losses.   


Professional macroeconomic forecasts, such as those from central banks, usually stem from both quantitative predictions and the judgment of economic experts \citep{Matsypura.2018}. For example, the German Bundesbank combines factor models, demand- and supply-side bridge equations and time series models for industrial production. Expert knowledge is then applied to these quantitative forecasts to determine the final projections for key macroeconomic indicators.\footnote{German Bundesbank. \emph{Forecasting models in short-term business cycle analysis: A workshop report}. URL: \url{https://www.bundesbank.de/Redaktion/EN/Downloads/Publications/Monthly_Report_Articles/2013/2013_09_forecasting.pdf}, last accessed on January~12, 2018.} The European Central Bank has a similar process and their projections \emph{\textquote{may incorporate a fair amount of judgment}} \citep{ECB.2016}. Their questionnaire concludes that \emph{\textquote{respondents consider their predictions to be \SI{40}{\percent} judgment-based}}.\footnote{European Central Bank. \emph{Results of a special questionnaire for participants in the ECB Survey of Professional Forecasters}. URL: \url{https://www.ecb.europa.eu/stats/prices/indic/forecast/shared/files/quest_summary.pdf}, last accessed on January~12, 2018.} Hence, it is likely that these forecasts 


Numerous studies have analyzed professional forecasts for their predictive performance and ability to identify potential biases \citep{Blanc.2015,Mostard.2011}. In the field of macroeconomic forecasting, a recent study \citep{Dovern.2011} finds that the distribution of a forecasts accuracy varies significantly across indicators, forecasters and nations. A further study reveals that consensus estimates are poor predictors of gross domestic product as compared with statistical models \citep{Jansen.2016}. However, the consensus estimates tend to perform better in periods of crisis. An intriguing approach would be to combine expert estimations and statistical forecasts \citep{Lessmann.2012}. 


Predicting macroeconomic variables is difficult, requiring time-intensive collection of economic data, which, as a result, is often out of date. However, our news source provides free and real-time access. Additionally, time series models are excellent at picking up trends, but perform poorly in situations of volatility \citep{Jansen.2016}. In contrast, our approach can incorporate additional predictors besides trend, as it extracts qualitative information, as well as overall sentiments, regarding the economic outlook from financial news. At the same time, our semantic path model accounts for interpretable results and can thus be further assessed by decision-makers. 


Our innovative, news-based methodology adds to previous literature and entails a number of important implications for researchers, as well as for the public and private sectors, by providing long-term forecasts of the macroeconomic climate. Here big data is once again shifting the frontiers of research \citep{Mortenson.2015} and is likely to change the way in which operations adapt to economic developments. As demonstrated by our previously discussed evaluations, the system elicits a high predictive accuracy, when compared with a number of benchmark models. This contributes to a growing body of literature that replaces man-made forecasts with computerized and fine-tuned predictive models. Our proposed methodology could be extremely valuable for small enterprises or countries in which macroeconomic forecasts are scarce due to the associated costs. 



\section{Conclusion} 
\label{sec:conclusion}


{The volatile nature of the economy affects the operation of firms in key dimensions, such as prices for goods and services, financial risk, utilization of supply chains and customer demand. Hence, it is important for managers to obtain an accurate prognosis of developments in the economy in order to direct operations accordingly. Hitherto, macroeconomic forecasts have been dominated by expert opinions and rather simple time series models, while the big data era has created new opportunities to enhance the predictive power of macroeconomic forecasts based on unstructured data sources.}


This work contributes a news-based methodology for predicting macroeconomic indicators. First, we experiment with conventional models from machine learning in order to predict macroeconomic outcomes on the basis of word occurrences and historic lags. 
Back-testing shows that this method outperforms the benchmark models in an out-of-sample evaluation for various prediction horizons and macroeconomic indicators. Yet text-based predictions represent a challenging undertaking due to potential overfitting for a high-dimensional input. {Instead, we suggest an alternative approach whereby the words from different semantic categories are mapped onto latent structures as a form of feature engineering. This shrinks the feature space considerably, thereby achieving superior out-of-sample performance in multiple experiments. Beyond that, we propose semantically-structured variant of partial least squares that reaches a comparable accuracy, but while fulfilling the demand of practitioners in being fully interpretable.} This contributes to a greater understanding of how qualitative information can be used to make long-term predictions for key macroeconomic indicators. 


This paper opens avenues for future research in a number of directions. First, the proposed semantic path model can facilitate interpretable predictions in a variety of settings involving natural language. Second, the approach can be seamlessly extended to include alternative data sources, including additional news sources and time series. Third, one could weigh corporate disclosures according to their market capitalization, since larger firms presumably have a great impact on the general economy than smaller firms.



\section*{References}

\setlength{\bibsep}{0.15\baselineskip}
{\sloppy\small
\setlength{\bibsep}{3pt}
\bibliographystyle{model5-names-no-doi}
\bibliography{literature}
}







\end{document}